%% file: main.tex
\documentclass[11pt,a4paper]{article}
\usepackage[a4paper,margin=1in]{geometry} %
\usepackage{graphicx}
\usepackage{amssymb}
\usepackage{amsmath}
\usepackage{algorithm}
\usepackage{algorithmic}
\usepackage{soul}
\usepackage[normalem]{ulem}

\usepackage[utf8]{inputenc} %
\usepackage[T1]{fontenc}    %
\usepackage{hyperref}       %
\usepackage{url}            %
\usepackage{booktabs}       %
\usepackage{amsfonts}       %
\usepackage{nicefrac}       %
\usepackage{microtype}      %
\usepackage{xcolor}         %

\newcommand{\titlerunning}[1]{}
\newcommand{\authorrunning}[1]{}

\newcommand{\orcidID}[1]{}
\newcommand{\email}[1]{\href{mailto:#1}{#1}}

\usepackage{colortbl}
\usepackage{nccmath, mathtools}
\usepackage{wrapfig,lipsum,booktabs}
\usepackage{verbatim}
\usepackage{bm}

\usepackage{microtype}
\usepackage{graphicx}
\usepackage{booktabs}
\usepackage{hyperref}
\usepackage{thmtools,thm-restate}
\usepackage{xcolor}
\usepackage{amsfonts}
\usepackage{stmaryrd}

\usepackage{float}
\usepackage{hyperref}
\usepackage{pgffor}
\usepackage{subcaption}
\usepackage{units}
\usepackage{cleveref}

\usepackage{multirow}

\usepackage{xspace}
\newcommand{\ar}{AR\xspace} 

\newcommand{\Tconst}{\textsc{const}\xspace}
\newcommand{\Tpoly}{\textsc{linear}\xspace} %

\newcommand{\Tsin}{\textsc{sin}\xspace}
\newcommand{\Tcos}{\textsc{cos}\xspace}

\newcommand{\Tadd}{$+$\xspace}
\newcommand{\Tmul}{$\times$\xspace}
\newcommand{\Tdiv}{$\div$\xspace}

\newcommand{\Tsub}{$-$\xspace}
\newcommand{\Texp}{\textsc{exp}\xspace}
\newcommand{\Tlog}{\textsc{log}\xspace}
\newcommand{\Tsqrt}{\textsc{sqrt}\xspace}
\newcommand{\Tone}{\textsc{1.0}\xspace}

\DeclareMathOperator*{\argmax}{arg\,max}
\DeclareMathOperator*{\argmin}{arg\,min}

\newcommand{\constraint}{\varobslash}

\newcommand\RCOMMENT[1]{\hfill\(\triangleright\) #1}

\definecolor{darkgreen}{rgb}{0.0, 0.5, 0.0}
\colorlet{gp}{blue}
\colorlet{aif}{red}
\colorlet{lspt}{brown}
\colorlet{lm}{darkgreen}
\colorlet{dsr}{violet}

\newcounter{subroutine}
\makeatletter

\usepackage{amsthm}
\newtheorem{lemma}{Lemma}

\begin{document}

\title{Deep Symbolic Optimization: Reinforcement Learning for Symbolic Mathematics}
\author{
Conor F. Hayes \and
Felipe Leno Da Silva \and
Jiachen Yang \and
T. Nathan Mundhenk \and
Chak Shing Lee \and
Jacob F. Pettit \and
Claudio Santiago \and
Sookyung Kim \and
Joanne T. Kim \and
Ignacio Aravena Solis \and
Ruben Glatt \and
Andre R. Goncalves \and
Alexander Ladd \and
Ahmet Can Solak \and
Thomas Desautels \and
Daniel Faissol \and
Brenden K Petersen \and
Mikel Landajuela$^{*}$
}

\date{}

\maketitle

\begin{center}
Computational Engineering Division, Lawrence Livermore National Laboratory, USA\\
$^{*}$ Corresponding author. E-mail: (Mikel) \email{landajuelala1@llnl.gov}
\end{center}

\begin{abstract}

Deep Symbolic Optimization (DSO) is a novel computational framework that enables symbolic optimization for scientific discovery, particularly in applications involving the search for intricate symbolic structures. One notable example is equation discovery, which aims to automatically derive mathematical models expressed in symbolic form. In DSO, the discovery process is formulated as a sequential decision-making task. A generative neural network learns a probabilistic model over a vast space of candidate symbolic expressions, while reinforcement learning strategies guide the search toward the most promising regions. This approach integrates gradient-based optimization with evolutionary and local search techniques, and it incorporates in-situ constraints, domain-specific priors, and advanced policy optimization methods. The result is a robust framework capable of efficiently exploring extensive search spaces to identify interpretable and physically meaningful models. Extensive evaluations on benchmark problems have demonstrated that DSO achieves state-of-the-art performance in both accuracy and interpretability. In this chapter, we provide a comprehensive overview of the DSO framework and illustrate its transformative potential for automating symbolic optimization in scientific discovery.

\end{abstract}

\begin{center}
\end{center}

\tableofcontents

\input{introduction}
\input{dso}

\input{dso_training}
\input{experiments}

\input{conclusion}

\bibliographystyle{plain}
\bibliography{main}
\end{document}

%% file: introduction.tex
\section{Introduction}
Many challenging scientific problems can be formulated as discrete optimization problems, where the goal is to find the best solution from a combinatorial search space \cite{bhcombinatorial,bohacek1996art,erlanson2016twenty}. Due to the vast number of potential solutions (often reaching %
far beyond than the number of atoms in the observable universe%
), these problems are often intractable. In recent years, Deep Learning (DL) has been utilized to tackle these problems, and has been shown to enhance scientific discovery across many disciplines \cite{wang2023scientific,fawzi2022discovering,townshend2021geometric,Landajuela2022IntracardiacEI}. 

DL employs Neural Networks (NN) \cite{lecun2015deep,goodfellow2016deep} as function approximators that exhibit complex functional forms involving numerous nested non-linear operators and affine transformations \cite{rumelhart1985learning}. While the applications of DL have largely been successful \cite{jumper2021highly,bellemare2020autonomous}, this complexity poses a significant challenge to the deployment of DL models in real-world settings due to the difficulty to understand, verify, trust, and predict the underlying models' behaviour \cite{dayhoff2001artificial,tu1996advantages,london2019artificial}.

In contrast, the field of neuro-symbolic AI \cite{hitzler2022neuro,dingli2023neuro} aims to combine the strengths of both DL and traditional symbolic AI by letting the NN manipulate symbols and reason about them. At the core of many neuro-symbolic approaches, there is a
combinatorial optimization problem involving the \emph{assembly of structures composed of symbols}. We refer to this class of problems as \emph{Symbolic Optimization} (SO) \cite{barmpalexis2011symbolic}. Symbolic constructs can facilitate human understanding, while also being transparent and exhibiting predictable behaviour. For example, SO has been used for equation discovery \cite{tenachi2023deep}, a problem also known as \emph{Symbolic Regression} (SR), where the goal is to construct mathematical expressions that fit a dataset \cite{schmidt2009distilling,sahoo2018learning,kusner2017grammar}. 
SO has also been used for policy search \cite{landajueladiscovering} for control tasks.
From the learning perspective, the use of symbolic operators can be regarded as enforcing a strong form of regularization, one that constrains the learned model to be expressed in a succinct symbolic form.

In this chapter, we describe in detail a framework that binds both DL and SO together. Our framework leverages the representational capacity of NNs to generate interpretable expressions, while entirely bypassing the need to interpret the network itself. We propose \emph{Deep Symbolic Optimization} (DSO)\footnote{All of our implementations are available at \url{https://github.com/dso-org/deep-symbolic-optimization}.}, a gradient-based approach for SO based on Reinforcement Learning (RL). DSO utilizes neural-guided search, whereby an autoregressive generative model emits a distribution over tokens. Solutions are sampled from the distribution, instantiated, and evaluated based on their fitness or reward \footnote{The reward function can be defined with respect to any property of the symbolic arrangement, not just fitness.}. The fitness is then used as the signal from which the autoregressive model is trained. 

As we show, our DSO framework is modular and can be enhanced by adding various components. 
In \cref{sec:dso}, we present the core DSO framework, which consists of an autoregressive model, and outline how in-situ priors and constraints can be used to leverage domain knowledge within the DSO framework. In \cref{sec:training_dso}, we show how different training strategies, can be used to update of the autoregressive model. We show how imitation learning from expert data via priority queue training and genetic programming population seeding can be integrated with the DSO framework to boost performance. We also present a methodology for integrating large-scale pre-training into the DSO framework. In \cref{sec:experiments}, we evaluate each presented component using established symbolic regression benchmarks from the literature. As a final evaluation, we combine the various DSO components into a Unified DSO (uDSO) framework, and evaluate uDSO against contemporary algorithms on SR benchmarks. Using uDSO, we perform extensive ablations to determine the contributions of each component in overall performance.

%% file: dso.tex
\section{The Deep Symbolic Optimization Framework}
\label{sec:dso}
\begin{figure}[h]
    \centering
    \includegraphics[scale=0.55]{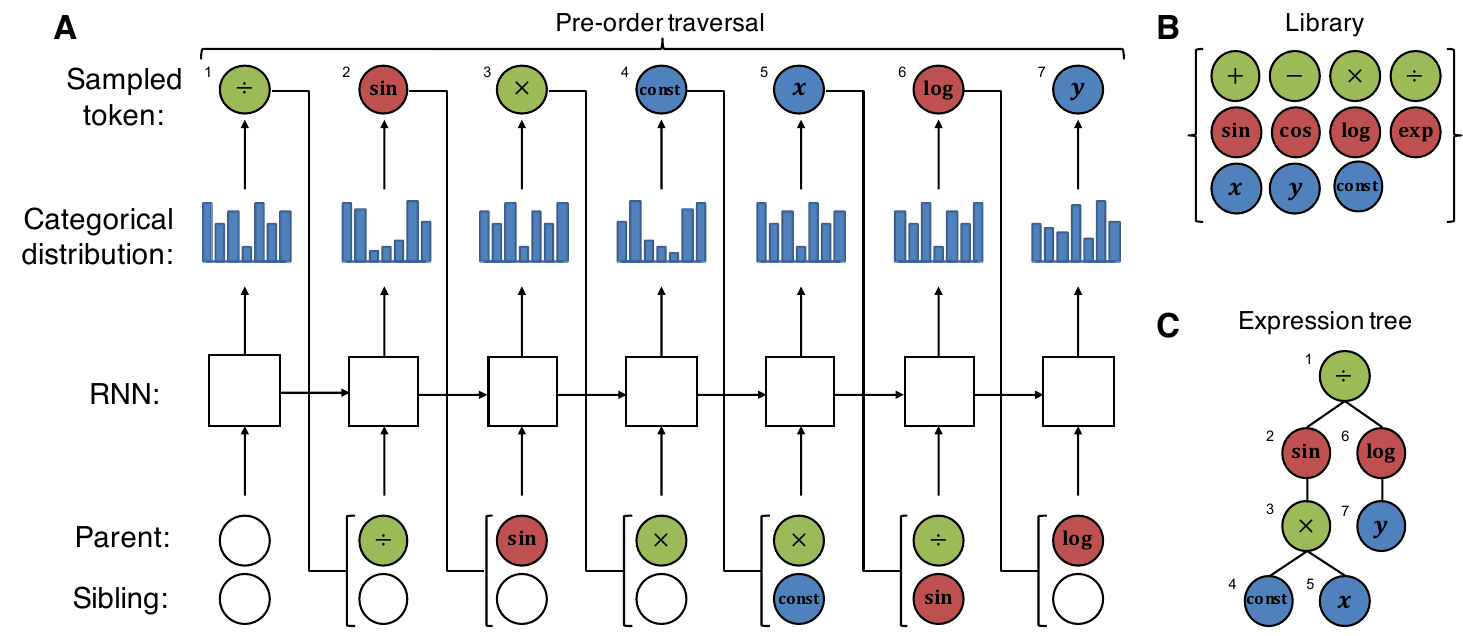}
    \caption{\textbf{DSO overview}. \textbf{A}. Example of sampling an expression from the generative model. In the figure, we use a Recurrent Neural Network (RNN).
  For each token, the RNN emits a categorical distribution over tokens, a token is sampled, and the parent and sibling of the next token are used as the next input to the RNN.
  In this example, the sampled expression is $\sin(cx)\//\log(y)$, where the value of the constant $c$ is optimized with respect to an input dataset. Starting at the root node, a token is sampled from the emitted categorical distribution. Subsequent tokens are sampled autoregressively until the tree is complete (i.e., all tree branches reach terminal nodes). The resulting sequence of tokens is the tree's pre-order traversal, which can be used to reconstruct the tree and instantiate its corresponding expression. Colors correspond to the number of children for each token. White circles represent empty tokens. \textbf{B}. The library of tokens. \textbf{C}. The expression tree sampled in \textbf{A}. Figure reproduced from \cite{petersen2019deep}.}
    \label{fig:dso}
\end{figure}

In this section, we present Deep Symbolic Optimization (DSO), a general framework for symbolic optimization in scientific problems.
DSO treats the discovery process as a sequential-decision making process where a probabilistic generative model is learned over the space of possible solutions. We describe the core components of the DSO framework below.
See \cref{fig:dso} for an overview of the method.

\subsection{Defining the Discovery Problem}
\label{sec:problem_formulation}

DSO is a general framework for solving discrete optimization problems. However, it is particularly well-suited for problems that exhibit the following characteristics:
\begin{enumerate}
    \item The fitness of a solution can be evaluated using a black-box reward function.
    \item The solution can be represented as a variable-length sequence of discrete tokens.
    \item The solution presents a sequential structure with prefix-dependent positional constraints.
\end{enumerate}
Traditional optimization methods, such as integer programming or evolutionary algorithms, can be inefficient or intractable for these types of problems. In this chapter, we focus on problems that involve search over mathematical expressions, which are common in scientific discovery and present the aforementioned characteristics.
Note however, that DSO has also been successfully applied to other domains, such as
policy search for control tasks \cite{landajueladiscovering},
power converter design \cite{glatt2021deep,lenoAutoTG} and drug discovery \cite{dasilva2023multi_fidelity,faris2024pareto}.

In mathematical applications, the search space is normally an abstract functional space $\mathcal{T}$ that is unbounded and infinite-dimensional. To allow for a computational representation of the search space, DSO encodes mathematical expressions as \emph{symbolic expression trees}. The expression tree is represented as a binary tree where mathematical operators are internal nodes and input variables (or constants) are terminal nodes. The mathematical operators and input variables are represented as \emph{tokens} from a predefined \emph{library} $\mathcal{L}$. To further simplify the search space, DSO uses the pre-order traversal of the expression tree (depth-first, and then left-to-right) to represent the expression as a sequence of tokens $\tau = \langle \tau_1, \ldots, \tau_n \rangle$, where $\tau_i \in \mathcal{L}$ and $n$ is the length of the sequence. See \cref{fig:dso} for an example of a library, an expression tree and its corresponding token sequence.

In the context of symbolic optimization, the objective is to find an optimal sequence of tokens $\tau$ that maximizes a given black-box reward function $R: \tau \rightarrow \mathbb{R}$. The solution of a symbolic optimization problem is given by:
\begin{equation}\label{eq:hard_problem}
      \argmax_{\tau \in \mathcal{L}^n, n \in \mathbb{N}}  R(\tau) \textrm{ with } \tau=  \langle \tau_1, \dots, \tau_n \rangle.
\end{equation}
The main challenge of this problem is searching the set of possible sequences, which are variable in length and have a combinatorial structure.

In the case of Symbolic Regression (SR), the goal is to find a mathematical expression that fits a given dataset $D = (X,y) = \{(x_1^{(i)}, \dots, x_d^{(i)}, y^{(i)})\}_{i=1}^N$. A common reward function is the inverse of the normalized mean squared error (NMSE) \cite{petersen2021deep} between the expression and the dataset., i.e, $R(\tau) = \frac{1}{1 + \text{NMSE}}$.
In SR we seek to illuminate the true underlying process that generated the data. Thus, the process of symbolic regression is analogous to how a physicist may derive a set of fundamental expressions to describe a natural process. SR has been shown to be NP-hard even for low-dimensional data \cite{virgolin2022symbolic}.

\subsection{Autoregressive Modeling for Combinatorial Search}
\label{sec:ar}

Deep learning techniques have successfully been applied to combinatorial optimization problems across many settings \cite{bengio2021machine,mazyavkina2021reinforcement,khalil2017learning,schuetz2022combinatorial}, where many of these methods utilize autoregressive models to generate solutions \cite{bello2016neural,biggio2021neural,kamienny2022end}.
Following this trend, DSO uses an Autoregressive (\ar) model with parameters $\theta$ to generate token sequences. To fix ideas, we describe DSO using a Recurrent Neural Network (RNN) as the \ar model in our figures and examples. However, a GPT like model \cite{radford2018improving} could also be used.

The \ar model is used to generate sequences as follows.
At position $i$, the \ar model produces a vector of logits $\psi^{(i)}$
conditioned on the previously generated tokens $\tau_{1:(i-1)}$, i.e., $\psi^{(i)} = \text{\ar}(\tau_{1:(i-1)}; \theta)$.
The new token $\tau_i$ is sampled from the distribution
\begin{equation}\label{eqn:cond_dso}
p(\tau_i | \tau_{1:(i-1)}, \theta) = \text{Softmax}(\psi^{(i)})_{\mathcal{L}(\tau_i)},
\end{equation}
where $\mathcal{L}(\tau_i)$ is the index in $\mathcal{L}$ corresponding to node value $\tau_i$.
This process is illustrated in \cref{fig:dso}.

\subsection{Integrating Priors and Constraints to Prune the Search Space}
\label{sec:prior_contraints}
\begin{figure}
    \centering
    \includegraphics[scale=0.6]{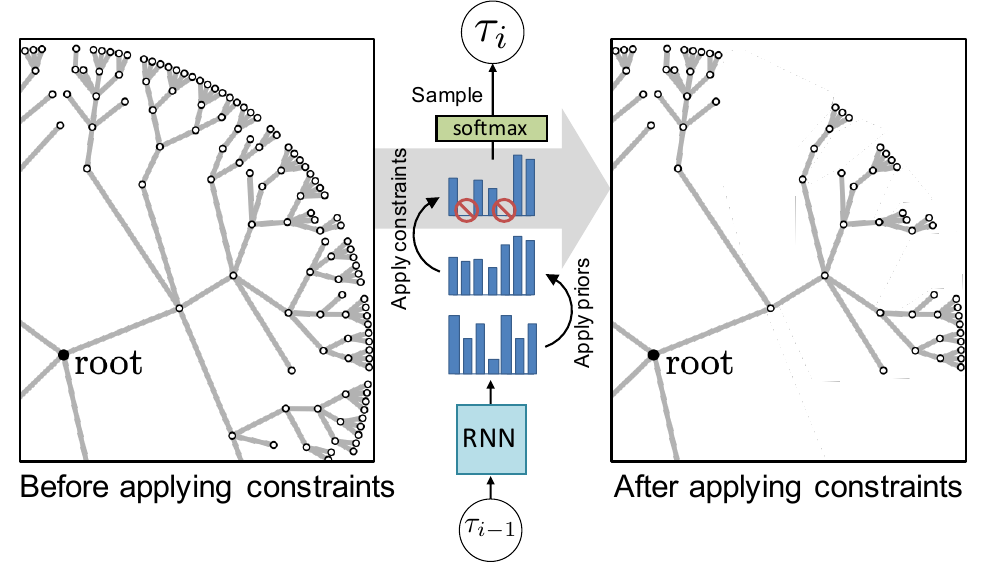}
    \caption{\textbf{Pruning the search tree in neural-guided search via in situ constraints.} The \ar model (RNN in this case) produces logits for each token in the library. Priors and constraints can be learned or user defined, and, are used to adjust the logits guiding the search towards well-performing sequences. The adjusted logits are used to sample the next token in the sequence. The process is repeated until the sequence is complete. Figure reproduced from \cite{petersen2021incorporating}.}
    \label{fig:dso_constraints}
\end{figure}

A key benefit of sampling tokens autoregressively is the opportunity to incorporate prior knowledge into the search phase at each step of the generative process. For problems with prefix-dependent positional 
structure or constraints, i.e., problems where certain tokens are more likely to follow others, or where certain tokens
can not follow others, the \ar model can be provided with well crafted priors and constraints to prune the search tree as outlined in \cref{fig:dso_constraints}. 

Both constraints and priors can be generally defined as \emph{logit adjustment vectors}, denoted $\psi_{\constraint}$ and $\psi_\circ$, which are added to the logits $\psi$ emitted by the \ar during sampling: 
\begin{equation}
    \tau_i \sim \textrm{Categorical}(\textrm{Softmax}(\psi^{(i)} + \psi^{(i)}_{\constraint} + \psi^{(i)}_\circ))_{\mathcal{L}(\tau_i)}.
    \label{eq:sampling}
\end{equation}
Priors \emph{bias} the search, but do not eliminate possible sequences or reduce the size of the search space. In contrast, constraints prune the search space by eliminating possible sequences. 

Priors can be used to bias the search by assigning higher or lower value to a given token(s) by altering the logits from the controller. A positive value encourages sampling for the respective tokens, whereas negative values dissuade sampling. For example, a prior could be assigning logits values in a way that the initial probability of sampling tokens of each arity is the same (avoiding oversampling tokens of a given arity if they are more common). A simple example of constraint is setting a maximum size for the sampled sequence. This constraint would assign zero value to all tokens until the maximum size is achieved, which then would result in negative infinity being assigned to all tokens but the terminal one.

Note that gradients with respect to \ar model parameters can be computed similarly in \cref{eq:sampling} as when using \cref{eqn:cond_dso}, since neither constraints or priors depend on the \ar. This allows to seamlessly combine this framework with gradient-based learning methods.

\subsection{The Challenge of Constant Optimization in Symbolic Expressions}
\label{sec:constants}

In physics and engineering problems, equations often include numeric constants, which may represent fundamental physical constants (e.g., the speed of light) or tunable parameters optimized to fit models to data.

To handle constants in the DSO framework, we introduce a special constant token in the library $\text{const}(\beta)$ that acts as a placeholder for a numeric value $\beta$. The \ar model can sample this token as any other token in the library, producing skeletons of expressions that include placeholders. To evaluate the fitness of an expression with constant tokens, we optimize the corresponding continuous values using a nonlinear optimization algorithm, e.g. BFGS. This optimization process is performed as part of the reward computation before performing each training step. Typically, constant fitting is the most computationally expensive step in trial-and-error symbolic regression systems. The final reward that the \ar model receives is the reward of the expression with the optimized constants.

\begin{figure}[h]
    \centering
    \includegraphics[scale=0.5]{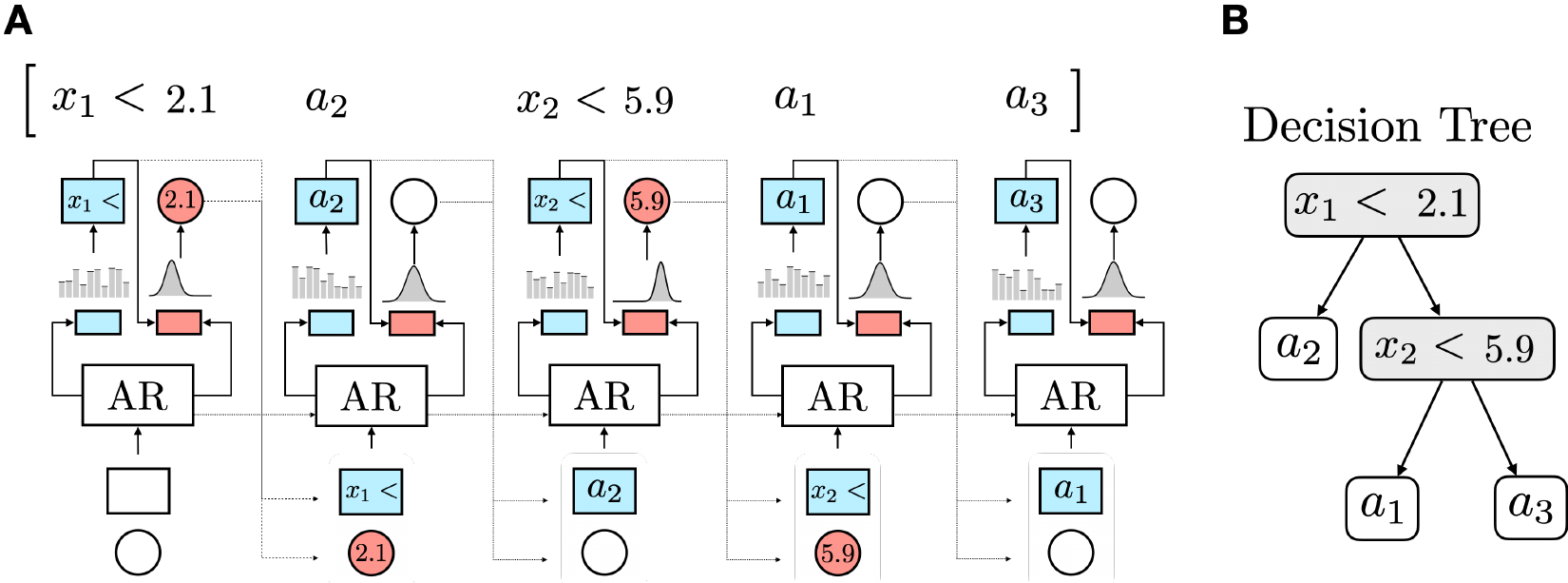}
    \caption{
    \textbf{DisCo-DSO overview} (application to decision tree search). 
    \textbf{A}. DisCo-DSO augments the autoregressive model with a continuous distribution over floating-point values that is conditioned on the discrete tokens. The observations are also augmented with the continuous values.
    \textbf{B}. Resulting decision tree sampled by DisCo-DSO. The continuous values are optimized jointly with the discrete tokens. Figure reproduced from \cite{pettit2024discodsocouplingdiscretecontinuous}. To be compared with \cref{fig:dso}.
    }
    \label{fig:disco_dso}
\end{figure}

Recently, the extension \emph{Discrete-Continuous Deep Symbolic Optimization} (DisCo-DSO) \cite{pettit2024discodsocouplingdiscretecontinuous} has been proposed to bypass the need for a separate optimization step for constants. DisCo-DSO extends the autoregressive model to jointly model a discrete distribution over symbolic tokens and a conditioned continuous distribution for floating-point values (see \cref{fig:disco_dso}). This unified approach is particularly valuable in mathematical modeling domains with many floating-point constants, such as decision tree learning.

%% file: dso_training.tex
\section{Training Strategies for Deep Symbolic Optimization}
\label{sec:training_dso}

In this section, we present a comprehensive overview of the training strategies that can be used to update the generative \ar model in the DSO framework. 
We explore a variety of training strategies, including RL, expert imitation learning, and large-scale pre-training. 

\begin{figure}
  \centering
  \begin{subfigure}[b]{0.3\textwidth}
    \centering
    \includegraphics[width=\textwidth]{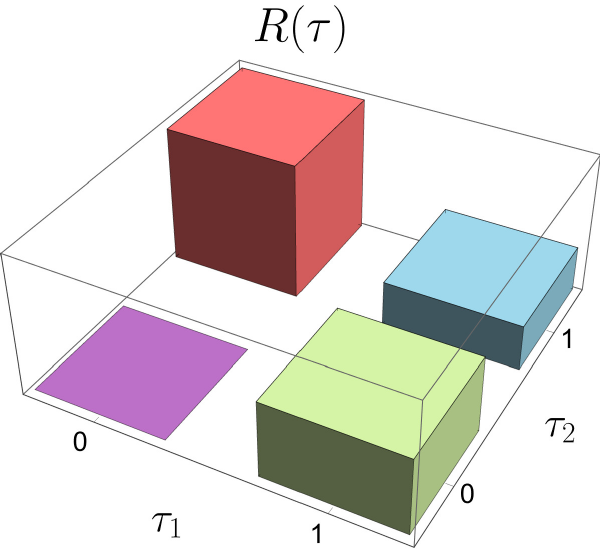}
    \caption{Discrete Objective Function}
  \end{subfigure}\hfill
  \begin{subfigure}[b]{0.3\textwidth}
    \centering
    \includegraphics[width=\textwidth]{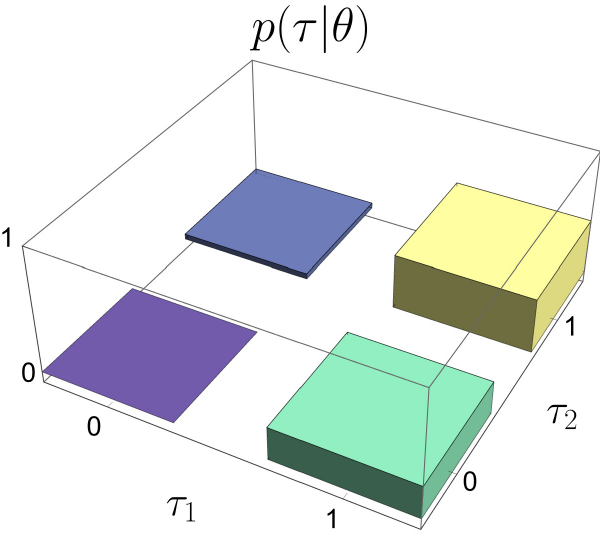}
    \caption{Initial Generative Model}
  \end{subfigure}\hfill
  \begin{subfigure}[b]{0.3\textwidth}
    \centering
    \includegraphics[width=\textwidth]{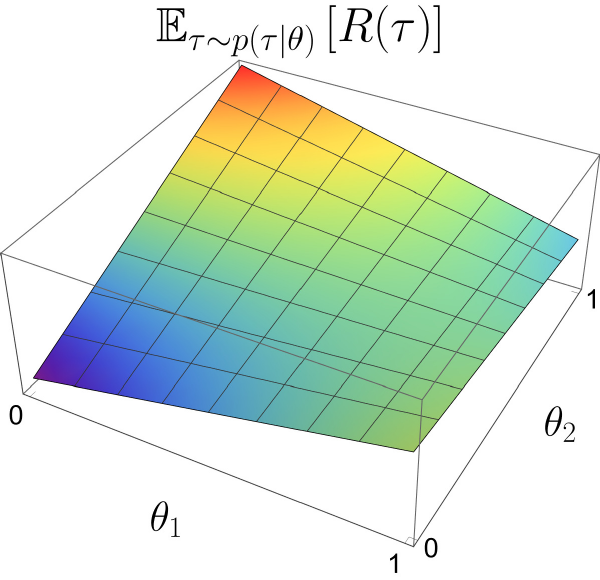}
    \caption{Expected Reward Relaxation}
  \end{subfigure}

  \vspace{1em} %

  \begin{subfigure}[b]{0.3\textwidth}
    \centering
    \includegraphics[width=\textwidth]{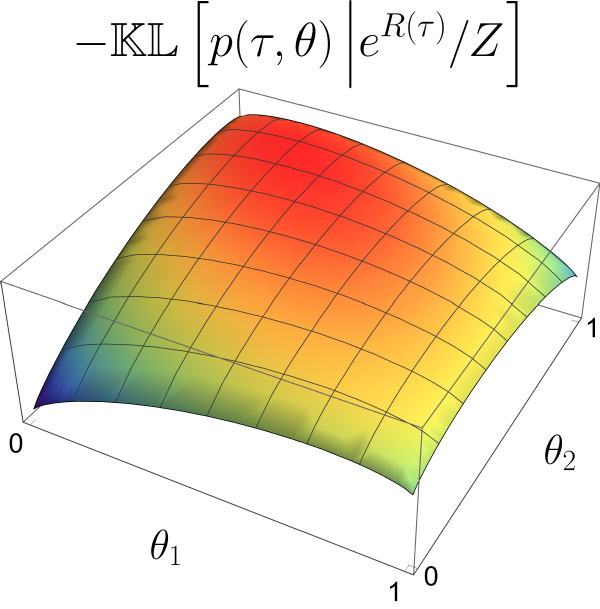}
    \caption{Entropy-Regularized Objective}
  \end{subfigure}\hfill
  \begin{subfigure}[b]{0.3\textwidth}
    \centering
    \includegraphics[width=\textwidth]{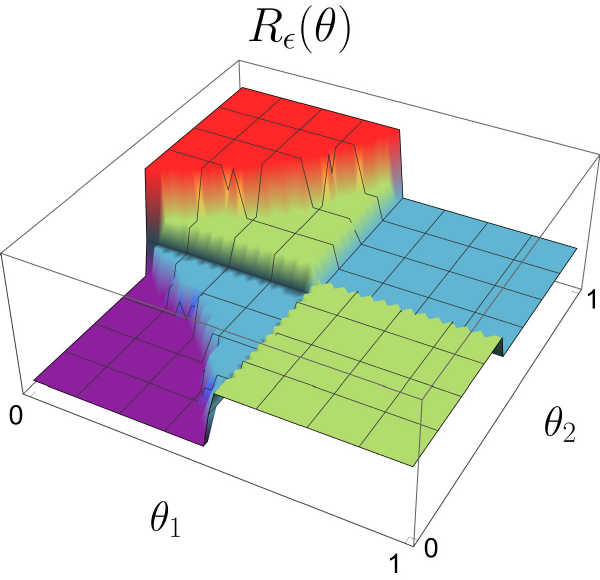}
    \caption{Reward Quantile Constraint}
  \end{subfigure}\hfill
  \begin{subfigure}[b]{0.3\textwidth}
    \centering
    \includegraphics[width=\textwidth]{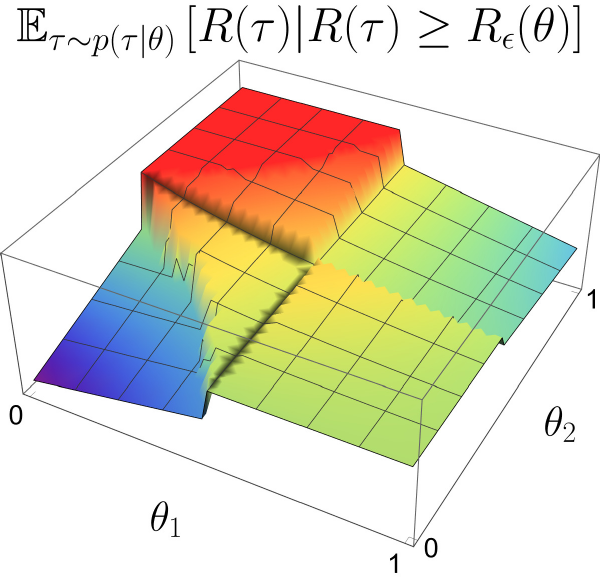}
    \caption{Quantile-Conditioned Expected Reward}
  \end{subfigure}

  \caption{%
    \textbf{Relaxation and Dual Formulation of the Discrete Optimization Problem.} 
    This figure illustrates the dual formulation of a discrete optimization problem in a toy example where the search space consists of four possible solutions: $\mathcal{T}=\{(0,0),(1,0),(0,1),(1,1)\}$. 
    (a) The black-box discrete objective function $R(\tau)$ assigns rewards to each point.
    (b) The generative model $p(\tau|\theta)=\mathcal{B}(\tau_1|\theta_1)\,\mathcal{B}(\tau_2|\theta_2)$ with $\theta_1=0.8,\theta_2=0.2$.
    (c) Relaxation via expected reward under the model.
    (d) Entropy regularization (negative prior divergence).
    (e) $(1-\epsilon)$-quantile constraint on the reward.
    (f) Expectation conditioned on that quantile.
  }
  \label{fig:dual_problem}
\end{figure}

The main idea is to reformulate the discrete optimization problem \cref{eq:hard_problem} as a continuous optimization problem, for which we can use gradient-based optimization techniques. 
See \cref{fig:dual_problem} for an illustration of this reformulation. 
The following lemma provides a theoretical justification.

\begin{lemma} (Dual Problem)
Assume that for all $\tau \in \mathcal{T}$, there exists $\theta_{\tau} \in \mathbb{R}^M$ such that $p(\tau | \theta_{\tau})$ is the Dirac delta distribution $\delta_{\tau}$.
Assume also that there exists a unique global maximum $\tau^{\star} = \argmax_{\tau \in \mathcal{L}^n, n \in \mathbb{N}} \left[ R( \tau )\right] $.
Consider $\tau^{\star}_p$ defined as,
\begin{equation}\label{eq:soft_problem}
\begin{cases}
    \theta^{\star}  = \argmax_{\theta \in \mathbb{R}^M} \mathbb{E}_{\tau \sim p(\tau | \theta)  } \left[ R( \tau ) \right],\\
    \tau^{\star}_{p} = \argmax_{\tau} p(\tau | \theta^{\star}).
\end{cases}
\end{equation}
Then, we have that $\tau^{\star} = \tau^{\star}_p$, i.e., 
the optimization problem (\cref{eq:hard_problem}) can be reformulated as a continuous optimization 
problem over the parameters $\theta$ of the generative model. 
\end{lemma}

\begin{proof}
Consider $\theta_{\tau^{\star}}$, i.e., the parameters of the generative model such that $p(\tau | \theta_{\tau^{\star}}) = \delta_{\tau^{\star}}$. 
We have that 
$$\mathbb{E}_{\tau \sim p(\tau | \theta)} \left[ R( \tau ) \right] \leq R(\tau^{\star}) = \mathbb{E}_{\tau \sim p(\tau | \theta_{\tau^{\star}} )} \left[ R( \tau ) \right], \ \forall \theta \in \mathbb{R}^M.$$
Therefore, $\theta^{\star} = \theta_{\tau^{\star}}$.
In consequence, $$\tau^{\star}_p 
= \argmax_{\tau} p(\tau | \theta^{\star} = \theta_{\tau^{\star}})
= \argmax_{\tau} \delta_{\tau^{\star}} 
= \tau^{\star}.$$
\end{proof}

Note that the optimization problem in \cref{eq:soft_problem} is a relaxation of the original problem in \cref{eq:hard_problem}. In practice, the generative model might not be able to represent the Dirac delta distribution, and the expectation and mode computation in \cref{eq:soft_problem} are typically approximated using Monte Carlo sampling. However, as we show extensively in \cref{sec:experiments}, the reformulation in \cref{eq:soft_problem} is effective in practice.

It is common to add an entropy term in \cref{eq:soft_problem} to encourage the generative model to explore the search space more effectively. The entropy term can be used to prevent the model from converging to a suboptimal solution. The entropy term is defined as 
$\mathbb{H}[p(\tau | \theta)] = -\mathbb{E}_{\tau \sim p(\tau | \theta)}[\log p(\tau | \theta)].$
The following lemma shows that, with the entropy term, the optimization problem in \cref{eq:soft_problem} is equivalent to minimizing the Kullback-Leibler ($\mathbb{K}\mathbb{L}$) divergence between the distribution $p(\tau | \theta)$ and a prior Gibbs distribution $\pi_0(\tau) \propto e^{R(\tau)}$.

\begin{lemma} (Entropy Regularization)
Consider the Gibbs distribution $\pi_0(\tau) = e^{R(\tau)}/Z$ as the prior distribution. Then, we have the following equivalence:
    \begin{equation}\label{eq:entropy_regularization}
        \argmin_{\theta} \mathbb{K}\mathbb{L}\left[
        p(\tau | \theta) \left| \pi_0(\tau) \right.\right] = 
        \argmax_{\theta} \left( \mathbb{H}[p(\tau | \theta)] + \mathbb{E}_{\tau \sim p(\tau | \theta)}[R(\tau)] \right).
    \end{equation}
\end{lemma}

\begin{proof}
Consider the Kullback-Leibler divergence between the distribution $p(\tau | \theta)$ and the prior distribution $\pi_0(\tau)$:
\begin{align*}
& \mathbb{K}\mathbb{L}\left[ p(\tau ,\theta )\left|\pi _0(\tau )\right.\right]=\int p(\tau ,\theta ) \log (p(\tau ,\theta )) \, d\tau -\int \log \left(\pi _0(\tau )\right) p(\tau ,\theta ) \, d\tau \\
& =-\mathbb{H}\left[p(\tau ,\theta )\right]-\int p(\tau ,\theta ) \log \left(\frac{e^{R(\tau )}}{Z}\right) \, d\tau  \\ 
& = -\mathbb{H}\left[p(\tau ,\theta )\right]-\int p(\tau ,\theta ) (R(\tau )-\log (Z)) \, d\tau \\
& =-\mathbb{H}\left[p(\tau ,\theta )\right]-\mathbb{E}_{p(\tau ,\theta )}[R(\tau)]+\log (Z).
\end{align*}
Note that the term $\log (Z)$ is a constant and does not depend on $\theta$. 
Therefore, minimizing the Kullback-Leibler divergence is equivalent to maximizing the entropy of the distribution $p(\tau | \theta)$ plus the expected reward.
\end{proof}

See \cref{fig:dual_problem}~(d) for an illustration of the entropy regularization problem.
The minimization of the Kullback-Leibler divergence of the distribution $p(\tau | \theta)$ and the prior distribution $\pi_0(\tau) \propto e^{R(\tau)}$ has been used in other 
works like the Generative Flow Networks literature \cite{bengio2023gflownet,shen2023towards} to promote exploration and avoid computing the partition function $Z$.
In the following sections, we explore different strategies to optimize the generative model in DSO against the objective functions in \cref{eq:soft_problem} and \cref{eq:entropy_regularization}.

A final reformulation of the optimization problem in \cref{eq:soft_problem} is to maximize the best-case performance of the generative model.
To that end, we define $R_{\epsilon}(\theta)$ as the $1 - \epsilon$ quantile of the distribution of rewards under the current generative model $p(\tau | \theta)$. This quantity is shown in \cref{fig:dual_problem}~(e). We then consider a new learning objective using the quantile conditioned reward expectation $\mathbb{E}_{\tau \sim p(\tau | \theta)} \left[ R(\tau) \mid R(\tau) \geq R_{\epsilon}(\theta) \right]$. 
This objective is illustrated in \cref{fig:dual_problem}~(f). This objective aims to increase the reward of the top $\epsilon$ fraction of samples from the distribution, without regard for the other samples below the $\epsilon$ threshold. The policy gradient of this new learning objective is discussed in the following sections.

\subsection{Reinforcement Learning for Sequence Generation}
\label{sec:rl}

The readers familiar with RL will recognize that the objectives in \cref{eq:soft_problem} and \cref{eq:entropy_regularization} are standard in the RL literature. In RL, the reward function is normally a black-box, and the goal is to maximize the expected reward. The entropy term is used to encourage exploration and prevent the model from converging to a suboptimal solution \cite{haarnoja2018soft}. Many RL methods have been developed to optimize for this setting for problems of varying complexity and size \cite{silver2018general,fawzi2022discovering,watkins1992q}. Consequentially, RL 
techniques are a natural starting point for training the \ar model under DSO.  

Below, we discuss connections of the discrete optimization problem to Markov decision processes (MDPs) and the policy gradient algorithm \cite{sutton2018reinforcement}. We then introduce a novel risk-seeking policy gradient algorithm that aims to maximize the best-case performance of the generative model.

\subsubsection{Framing Discrete Optimization as a Markov Decision Process}
\label{sec:dso_rl}

In order to utilize RL objective functions for DSO training, we integrate DSO into the RL framework. RL models sequential decision making problems using a Markov decision process (MDP). Therefore, to connect DSO and RL, we frame DSO as a RL problem  with states, actions, and rewards.

\begin{itemize}
    \item \textbf{States}:  The observable state consists of the previously sampled tokens $\tau_{1:t}$. For many domains of interest for DSO, there is a hierarchical relation between the tokens. For example, for symbolic regression, if a $div$ token is sampled, the following tokens will form the operands for the division, and this happens recursively until the whole expression is formed. We leverage this hierarchical relation to represent sequences as \textit{symbolic trees}. Symbolic trees are helpful because the previously sampled token may be very distant from the next token to be sampled in the tree. For example, the fifth and sixth tokens sampled in \cref{fig:dso} are adjacent nodes in the traversal but are four edges apart in the expression tree. We therefore capture hierarchical information from the symbolic tree and use it as agent observations in the form of the parent and sibling nodes of the token being sampled.    

\item \textbf{Actions}: Actions correspond to tokens from which the \ar model can sample. Therefore, the set of actions correspond to the library.

\item \textbf{Reward}: As previously mentioned the fitness of generated sequences are evaluated using a reward function, from which learning can take place. Partial sequence cannot be evaluated, therefore the reward function is only used to evaluate complete sequences, and a reward of zero, $r = 0$, is received at each timestep when sampling. At the final timestep (when a complete sequence has been generated) an undiscounted terminal reward $r = \mathcal{R}(\tau)$ is received.

\end{itemize}

\subsubsection{Policy Gradient Methods}
\label{sec:pg}
To train the \ar model we first consider the standard policy gradient objective. The goal is to maximize $J_{std}$, defined as the expectation of a reward function $R(\tau)$ under expressions from the distribution: $J_{std} = \mathbb{E}_{\tau \sim p(\tau | \theta)}\left[ R(\tau) \right]$.
The standard REINFORCE policy gradient \cite{williams1992simple} can be used to maximize this expectation via gradient ascent:
\begin{equation}
\nabla_\theta J_{std}  = \nabla_\theta \mathbb{E}_{\tau \sim p(\tau | \theta)}\left[ R(\tau) \right]
= \mathbb{E}_{\tau \sim p(\tau | \theta)}\left[ R(\tau) \nabla_\theta \log p(\tau | \theta) \right].
\label{eqn:pg}
\end{equation}
This result allows one to estimate the expectation using samples from the distribution. Specifically, an unbiased estimate of $\nabla_\theta J_{std}$ can be obtained by computing the sample mean over a batch of $N$ sampled expressions $\mathcal{T} = \{\tau^{(i)}\}_{i=1}^N$:
\begin{equation*}
\nabla_\theta J_{std} \approx \frac{1}{N} \sum_{i=1}^N R(\tau^{(i)}) \nabla_\theta \log p(\tau^{(i)} | \theta).
\end{equation*}
In practice this unbiased gradient estimate has high variance. To reduce variance, it is common to subtract a baseline function $b$
from the reward.
As long as the baseline is not a function of the current batch of expressions, the gradient estimate is still unbiased.
Common choices of baseline functions are a moving average of rewards or an estimate of the value function.

\subsubsection{Risk-Seeking Policy Gradient Approaches}
\label{sec:rspg}
The traditional policy gradient objective, presented in \cref{eqn:pg}, is only sufficient for problem domains where the goal is to maximize the average performance of a policy, e.g., control tasks \cite{brockman2016openai}. In some settings, performance may be measured by the single or few best-performing samples found during training, e.g., symbolic regression or neural architecture search \cite{zoph2017neural}. As a result, the policy gradient objective is not appropriate for these domains given there is a contradiction between the learning objective and the performance measure, also known as the \emph{expectation problem}. To address this disconnect, we propose a risk-seeking policy gradient algorithm that focuses on \emph{maximizing best-case performance} and trains the \ar model to find better fitting expressions.

We define $R_{\epsilon}(\theta)$ as the $1 - \epsilon$ quantile of the distribution of rewards under the current policy. Using $R_{\epsilon}(\theta)$, we define a new learning objective, $J_{risk}(\theta; \epsilon)$, parameterized by $\epsilon$:
\begin{equation*}
    J_{risk}(\theta; \epsilon) = \mathbb{E}_{\tau \sim p(\tau | \theta)} \left[ R(\tau) \mid R(\tau) \geq R_{\epsilon}(\theta) \right].
\end{equation*}
The defined objective aims to increase the reward of the top $\epsilon$ fraction of samples from the distribution, without regard for the other samples below the $\epsilon$ threshold. $J_{risk}(\theta; \epsilon)$ is therefore a \emph{risk-seeking} policy gradient that aims to maximise the best-case performance at the expense of worst-case and average performance. The policy gradient of $J_{risk}(\theta; \epsilon)$ is given by:
\begin{equation*}
    \nabla J_{risk}(\theta; \epsilon) = \mathbb{E}_{\tau \sim p(\tau \mid \theta)} \left[ (R(\tau) - R_{\epsilon}(\theta))   \nabla_{\theta}  \log (p(\tau \mid \theta)) \mid \right (R(\tau) > R_{\epsilon}(\theta)].
\end{equation*}
Therefore, a Monte Carlo estimate of the gradient from a batch of $N$ samples can be computed as follows:
\begin{equation}
    \nabla J_{risk}(\theta; \epsilon) \approx \frac{1}{\epsilon N} \sum_{i}^{N} \left[ R(\tau^{i}) - R_{\epsilon}(\theta) \right] \cdot \mathbf{1}_{R(\tau^{i}) \geq R_{\epsilon}(\theta)} \nabla_{\theta} \log p(\tau^{(i)} \mid \theta),
\label{eqn:rspg}
\end{equation}
where $R_{\epsilon}(\theta)$ is the empirical $(1 - \epsilon)$ quantile of the batch of rewards, and $\mathbf{1}_x$ returns $1$ if condition $x$ is true and $0$ otherwise. Therefore, we have new learning objective which is similar to the Monte Carlo estimate of REINFORCE, however, with two key differences: (1) the theory suggests a specific baseline, $R_{\epsilon}(\theta)$, whereas the baseline for policy gradient methods in non-specific and selected by the user; (2) effectively only the top $\epsilon$ fraction of samples from each batch are used in the gradient computation. It is important to note that \cref{eqn:rspg} is general and can be applied to any policy gradient algorithm trained on batches, e.g., proximal policy optimization \cite{schulman2017proximal}. Additionally, the risk-seeking policy gradient is closely related to the EPOpt-$\epsilon$ algorithm used for robust reinforcement learning, which is based on a risk-averse policy gradient formulation \cite{tamar2014policy}. Finally, see \cite{landajuela2021improving} for further improvements to the risk-seeking policy gradient algorithm.

In \cref{alg:dsr}, we present the core DSO algorithm with the risk-seeking policy gradient. The different learning strategies presented in the following sections can be readily integrated into \cref{alg:dsr} by modifying the initial weights of the \ar model
(large scale pre-training), the sample generation process (imitation learning from expert data), or the gradient computation (maximum likelihood training using priority queue training).

\begin{algorithm*}
\caption{DSO with risk-seeking policy gradient}
\label{alg:dsr}
$\textbf{input}$ learning rate $\alpha$; entropy coefficient $\lambda_\mathcal{H}$; risk factor $\epsilon$; batch size $N$; reward function $R$ \\
$\textbf{output}$ Best fitting expression $\tau^\star$
\begin{algorithmic}[1]
\STATE Initialize \ar model with parameters $\theta$, defining distribution over expressions $p(\cdot | \theta)$
\STATE \textbf{repeat}
\begin{ALC@rpt}
\STATE $\mathcal{T} \leftarrow \{\tau^{(i)} \sim p(\cdot | \theta)\}_{i=1}^N$ \RCOMMENT{Sample batch of $N$ expressions}
\STATE $\mathcal{T} \leftarrow \{\textrm{OptimizeConstants}(\tau^{(i)}, R)\}_{i=1}^N$ \RCOMMENT{Optimize constants w.r.t. reward function}
\STATE $\mathcal{R} \leftarrow \{R(\tau^{(i)}) \}_{i=1}^N$ \RCOMMENT{Compute rewards}
\STATE $R_\epsilon \leftarrow (1 - \epsilon)$-quantile of $\mathcal{R}$ \RCOMMENT{Compute reward threshold}
\STATE $\mathcal{T} \leftarrow \{\tau^{(i)} : R(\tau^{(i)}) \geq R_\epsilon \}$ \RCOMMENT{Select subset of expressions above threshold}
\STATE $\mathcal{R} \leftarrow \{R(\tau^{(i)}) : R(\tau^{(i)}) \geq R_\epsilon \}$ \RCOMMENT{Select corresponding subset of rewards}
\STATE $\hat{g_1} \leftarrow \textrm{ReduceMean}( (\mathcal{R} - R_\epsilon) \nabla_\theta \log p(\mathcal{T} | \theta) )$ \RCOMMENT{Compute risk-seeking policy gradient}
\STATE $\hat{g_2} \leftarrow \textrm{ReduceMean}( -\lambda_\mathcal{H}\nabla_\theta \mathcal{H}(\mathcal{T} | \theta) )$ \RCOMMENT{Compute entropy gradient}
\STATE $\theta \leftarrow \theta + \alpha (\hat{g_1} + \hat{g_2})$ \RCOMMENT{Apply gradients}
  \IF{$\max\mathcal{R} > R(\tau^\star)$} 
  \STATE $\tau^\star \leftarrow \tau^{(\arg\max \mathcal{R})}$ \RCOMMENT{Update best expression}
  \ENDIF 
\end{ALC@rpt}
\STATE \textbf{return} $\tau^\star$
\end{algorithmic}
\end{algorithm*}

\subsection{Leveraging Expert Demonstrations through Imitation Learning}
\label{sec:expert_imitation}

In imitation learning an \emph{apprentice} learner attempts to solve a problem by mimicking the behaviour of an \emph{expert} learner \cite{anthony2017thinking}. By leveraging this idea, we use imitation learning to train the \ar model to imitate fit solutions (i) contained within a top-$K$ priority queue, and (ii) evolved from a genetic programming algorithm. 

\subsubsection{Priority Queue Training for Enhanced Learning}
\label{sec:pqt}
The parameters of an \ar model can be updated using an algorithm known as \emph{Priority Queue Training} (PQT) \cite{abolafia2018neural}, that utilizes imitation learning by maintaining a buffer of the (best) top-$K$ solutions. In this case the top-$K$ solutions are treated as the expert from which the apprentice \ar model can learn. During training the \ar model maximizes the probability of solutions within the $K$ best performing solutions. The priority queue is continually updated with higher performing solutions over time, where duplicates are discarded. 

On initialization the priority queue is empty, at each iteration the \ar model samples new solutions, and updates the priority queue with respect to the solutions fitness, keeping only the solutions which fall within the top-$K$ highest rewarded solutions. Following this approach, the PQT learning objective is defined as the maximum likelihood of the top-$K$ solutions:

\begin{equation*}
    J_{PQT}(\theta) = \frac{1}{K}\sum_{k=1}^{K}\log p(\tau^{(k)} \mid \theta).
\end{equation*}

By following this learning procedure, the \ar model (apprentice) and the priority queue (expert) bootstrap off of each other, whereby the \ar finds better solutions through exploration and the priority queue provides better training targets.

\subsubsection{Incorporating Genetic Programming}
\label{sec:gp}
It is also possible to define a Genetic Programming (GP) instance as an expert from which an apprentice can be trained. Both DSO and GP generate batches (or populations) of solutions \cite{koza1992genetic,petronski2018gprl}, however they do so using different methods. To generate a batch of solutions, DSO samples from an \ar model (with parameters $\theta$), where each batch, $\mathcal{T}_{\ar}$, is generated from scratch. In contrast, GP requires an existing population, which it can operate on, where generation $i$ of GP begins with a population of solutions $\mathcal{T}_{GP}^{(i)}$, and an application of one generation of GP yields an altered population $\mathcal{T}_{GP}^{(i+1)}$.

\begin{figure}[t]
    \centering
    \includegraphics[scale=0.55]{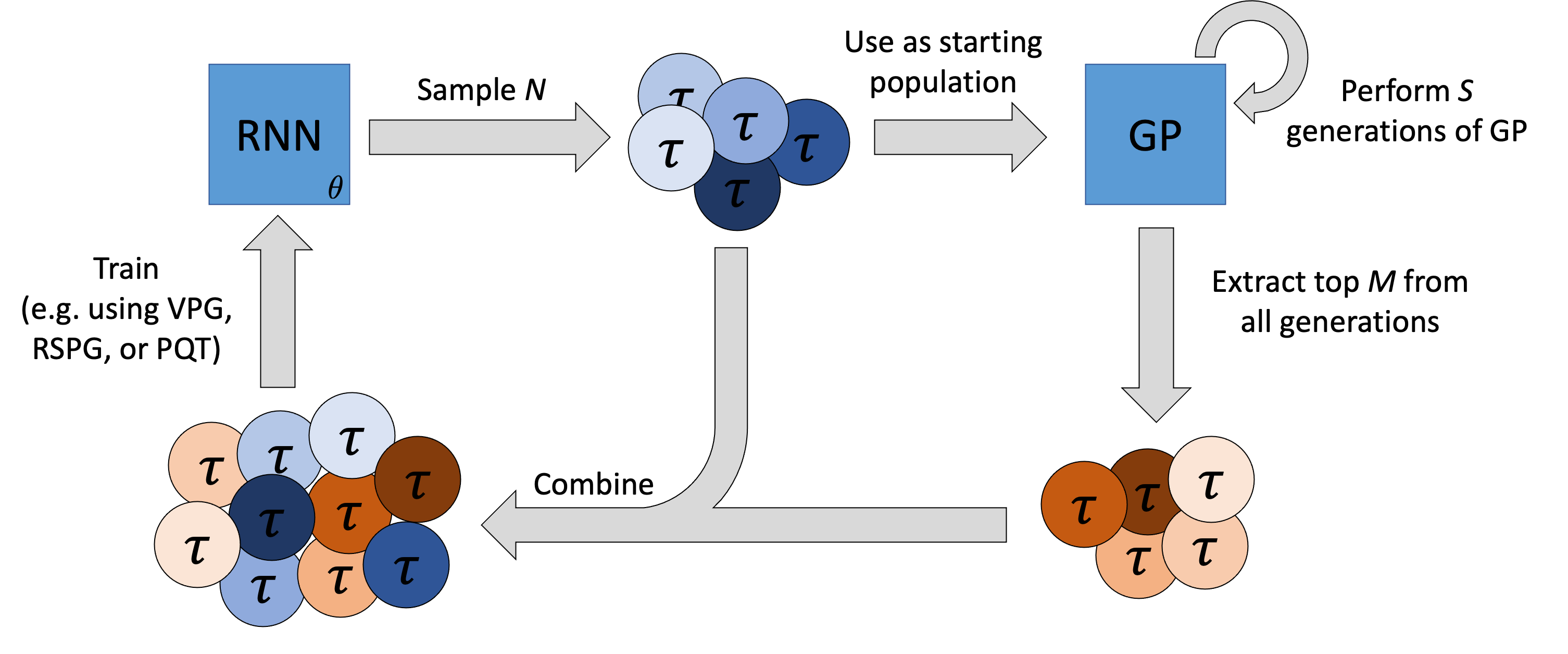}
    \caption{\textbf{Training DSO with genetic programming population seeding.} The AR model (a RNN in this case) generates $N$ samples, e.g. expression for symbolic regression. These samples are used as the start population for a GP component. GP then runs for $S$ generations. The top $M$ samples from GP are extracted, combined with the $N$ samples from the RNN, and are used to train the RNN. Since GP is stateless, it runs in a random restart-like fashion each time the RNN is sampled. Figure reproduced from \cite{mundhenk2021symbolic}.}
    \label{fig:dso_gp}
\end{figure}

We propose to use the most recent batch of solutions generated by the \ar model as the starting population for GP, $\mathcal{T}_{GP}^{(0)} = \mathcal{T}_{\ar}$. From here, we can perform $S$ generations of GP, resulting in a final GP population, $\mathcal{T}_{GP}^{S}$. Finally, we sub-select an elite set of top-performing samples, and include these samples in the gradient updates for the \ar model. The process constitutes once step of our algorithm, and is repeated until a maximum number of total solution evaluations is reached. Thus, GP acts as expert via an inner optimization loop where fit solutions are generated from which the \ar model can be trained. Similarly to PQT, the \ar model and GP bootstrap from each other. The GP evolves the population sampled by the \ar model, thus providing better training targets for DSO. Each step of our proposed optimization approach is illustrated in \cref{fig:dso_gp}. For all algorithmic and implementation details, we refer the reader to \cite{mundhenk2021symbolic}.

\subsection{Scaling Up: Large-Scale Pre-Training and Synthetic Demonstrations}
\label{sec:lspt}
\emph{Large-Scale Pre-Training} (LSPT) aims to tackle the problem of generalization by training a single model, that once trained can produce solutions for a given task in zero-shot manner. By incorporating pre-training with DSO, it is possible to take advantage of task specific pre-training to improve overall performance. Here, we explore model pre-training within the DSO framework.

\subsubsection{Initializing with a Pre-Trained Prior}
\label{sec:ptp}
Recall the DSO prior can be used to encode domain knowledge directly into the decision-making process. Therefore, this is a natural place to integrate pre-training into the DSO framework. A \emph{Pre-trained Prior} (PP) can directly inform the \ar model by adjusting the logits. In this case, we represent the prior using a NN. Given a sufficiently large dataset, we pre-train the NN prior to predict the next token in a sequence by minimizing the cross-entropy loss between the output of the PP and the given label. For example, mathematical expressions found in data sources, like Wikipedia, can be used to train a prior. This strategy is standard in natural language processing for training language models that predict the next word or character given a partial sequence.

To incorporate the PP into the generative process, we introduce a vector of logits $\psi$ computed by the PP:
\begin{equation*}
    (\psi^{(i)}_{PP}, c^{(i)}_{PP}) = \textrm{PP}(x, c^{(i-1)}_{PP}; \phi).
\end{equation*}
Finally, the logit vectors are added, a Softmax operator is applied, and the resulting probability vector defines a categorical distribution used to sample the next token $\tau_i$:
\begin{equation*}
    \tau_i \sim \textrm{Categorical}(\textrm{Softmax}(\psi^{(i)}_{DSO} + \lambda \psi^{(i)}_{(PP)} + \psi^{(i)}_\constraint)).
\end{equation*}
The hyperparameter $\lambda \in \mathbb{R}^+$ controls the strength of the PP.
We provide an interpretation of $\lambda$ based on the temperature of the Softmax function.
Note the Softmax function with temperature $T$ and logits $\psi$ is defined as: $\textrm{Softmax}_T(\psi) \circeq \textrm{Softmax}(\psi/T)$.
Multiplying logits by $\lambda$ yields:
\begin{equation*}
    \lambda \psi = \textrm{Inverse-Softmax}(\textrm{Softmax}(\lambda \psi)) = \textrm{Inverse-Softmax}(\textrm{Softmax}_{\lambda^{-1}}(\psi)),
\end{equation*}
where Inverse-Softmax is defined up to an arbitrary constant.
Thus, $\lambda$ can be viewed as the \textit{inverse temperature} of the PP's contribution to the final Softmax.
Higher temperatures (lower $\lambda$) result in a lower influence of the PP on the final Softmax. At the extreme, $\lambda \rightarrow 0$ corresponds to infinite temperature, in which case the PP is ignored. Note that the recurrent architecture used in the PP can be completely different than the one used in DSO; this allows the PP to be pre-trained offline. All details of this method can be found in the work of Kim et al. \cite{kim2021distilling}.

\subsubsection{Learning from Synthetic Demonstrations}
\label{sec:pretraining_encoder_decoder}
\begin{figure}[t]
    \centering
    \includegraphics[scale=0.5]{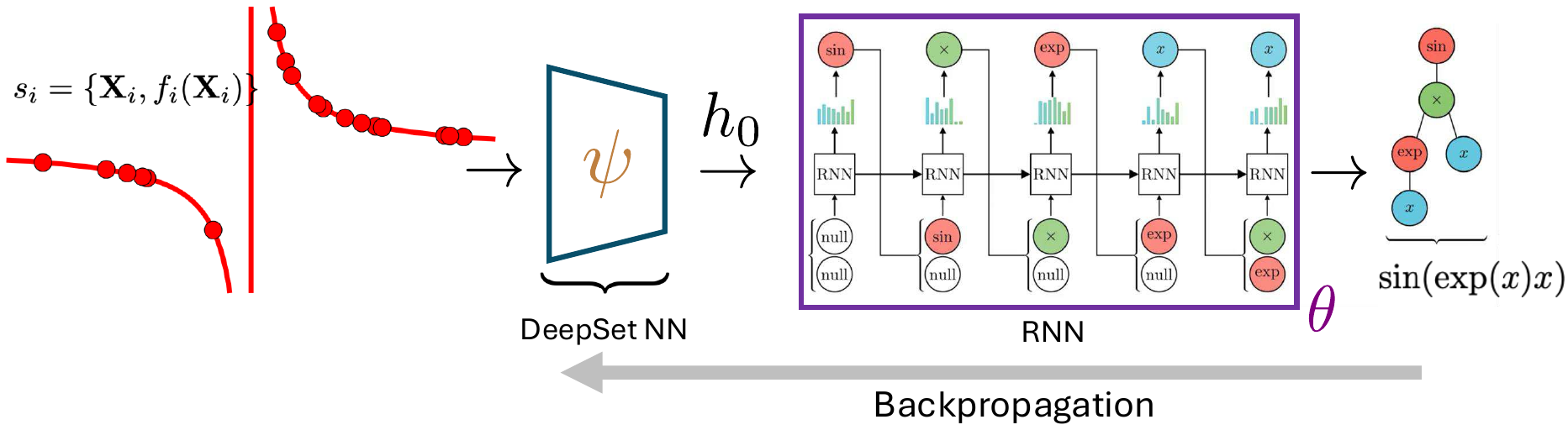}
    \caption{\textbf{Illustration of the pre-training process.} The DSO core \ar model is extended with a ST encoder (called DeepSet NN in the figure). The ST learns a latent representation $h_0$ of a dataset $(X, y)$. This representation is then passed as the initial state of the \ar model that decodes it into a solution $\tau$. The whole joint model is trained end-to-end on many problems using supervised learning, using ground truth synthetic demonstrations as labels.
    For more details, see \cite{landajuela2022unified}.}
    \label{fig:pretraining}
\end{figure}

To explicitly utilize pre-training for the generative model, we augment the \ar model to incorporate an encoder-decoder style architecture. Specifically, we use a \emph{Set Transformer} (ST) \cite{lee2019set} as the encoder, that learns a latent representation $h_{0}$ of a given dataset $(X, y)$. This representation is then passed as the initial state of the \ar model that decodes it into a solution $\tau$. This approach is utilized in \cite{landajuela2022unified} where a unified framework is proposed, see \cref{fig:pretraining} for an illustration of the resulting architecture.

The resulting architecture models the distribution $p(\tau|\theta, \mu, (X, y))$, where $\theta$ and $\mu$ are the parameters of the \ar model and ST, respectively. This model can be trained end-to-end on many problems using supervised learning, using synthetic solutions as labels, resulting in parameters $\theta^*$ and $\mu^*$. 
In the following, we assume that the synthetic demonstrations $(X,y)_g$ are generated from a generative model $p_{\mathcal{G}}$ that is different from the model $p(\tau|\theta, \mu)$ used for training.

In this work, we explore two different objectives for model pre-training. 
First, following  \cite{biggio2021neural}, we consider the Supervised Learning (SL) objective:
\begin{equation}
J_{\mathrm{SL}}(\theta, \mu ) :=
\mathbb{E}_{ \tau_g, (X, y)_g \sim p_{\mathcal{G}}}
\left [ \log  p \left( \tau_g |  \theta , \mu, (X, y)_g \right) \right].
\end{equation}

During fine-tuning, ground-truth labels will not be available.
To mitigate an objective mismatch between pre-training and fine-tuning objectives, we alternatively explore pre-training \textit{\`a la} RL \cite{bello2017neural}.
Specifically, we consider: 

\begin{equation}
\label{eq:vanilla_rl}
J_{\mathrm{RL}}(\theta, \mu) :=
\mathbb{E}_{ (X, y)_g \sim p_{\mathcal{G}} }
\left [ 
\mathbb{E}_{ \tau \sim p(\tau|\theta, \mu, (X,y)_g) }
\left [ 
R(\tau, (X,y)_g)
\right]
\right].
\end{equation}
However, this learning objective will maximize the expected reward. As previously mentioned there are many tasks in which best-case performance is a more accurate measure of success. As a result, we aim to pre-train $\theta$ and $\mu$ using the risk-seeking policy gradient objective function. Therefore, we extend the definition of risk-seeking policy gradient to optimize both $\theta$ and $\mu$ as follows:
\begin{equation}
    J_{risk}(\theta, \mu; \epsilon) = \mathbb{E}_{\tau \sim p(\tau | \theta)} \left[ R(\tau) \mid R(\tau) \geq R_{\epsilon}(\theta) \right].
\end{equation}

Despite extensive pre-training our algorithm may not be optimized for new out-of-distribution problems. To better optimize for such scenarios,  we propose to search for a symbolic expression by \textit{fine-tuning} (conditioned on the test problem) $p(\tau|\theta, \mu^\star, (X',y'))$ when given a new dataset $(X',y')$. The fine-tuning approach uses reinforcement learning, with initial conditions $\theta_0 = \theta^\star$. Therefore, during fine-tuning we only train the parameters of the decoder \ar model.

All algorithmic and implementation details for uDSO can be found in the work of Landajuela et al. \cite{landajuela2022unified}. 

%% file: experiments.tex
\section{Empirical Evaluation and Analysis}
\label{sec:experiments}
We evaluate DSO using the SR domain.   Here, we examine each aspect of DSO presented in \cref{sec:dso} and \cref{sec:training_dso}. For each experiment, we use a RNN as \ar model.

\subsection{Benchmark Datasets and Evaluation Protocols}
Symbolic regression is an active area of research with validated benchmark problems. Where applicable, we evaluate our proposed methods using the following benchmark datasets from the literature.

\subsubsection{Nguyen Symbolic Regression Benchmark Suite} In \cref{sec:exp:dso}, \cref{sec:exp:gp}, and \cref{sec:exp:lspt} we evaluate DSO using the task of symbolic regression using the Nguyen SR benchmark suite containing $12$ commonly used benchmark expressions developed by the symbolic regression community. Each benchmark is defined by a ground truth expression, a training and test dataset, and a set of allowed operators. The training data is used to compute the reward for each candidate expression, the test data is used to evaluate the best found candidate expression at the end of training, and the ground truth expression is used to determine whether the best found candidate expression was correctly recovered.

\subsubsection{SRBench}
In \cref{sec:exp:udso}, we use SRBench \cite{la2021contemporary}, an open-source and reproducible pipeline for benchmarking SR algorithms, as an evaluation benchmark. SRBench features 130 problems with hidden ground-truth analytic solutions and 122 real-world datasets with no known analytic model (``black-box'' problems) from the PMLB database \cite{olson2017pmlb}.

\subsection{Objective Function Analysis and Constraint Handling}
\label{sec:exp:dso}
To evaluate DSO, we trained the RNN using PQT, VPG, and RSPG (see \cref{sec:training_dso}). We also compared each DSO implementation against three strong symbolic regression baselines: (1) \textbf{GP} : a standard genetic programming based symbolic regression implementation, (2) \textbf{Eureqa} : a popular commercial software based on \cite{schmidt2009distilling}, and (3) \textbf{Wolfram} : commercial software based on Markov chain Monte Carlo and nonlinear regression. Finally, we evaluated DSO using the Nyguyen symbolic expression suite. For further evaluations and experimental details we refer the reader to \cite{petersen2021deep}.

\subsubsection{Constraint Handling Strategies}
For evaluation we impose the following in-situ constraints: \begin{enumerate}
    \item Expression are limited to a pre-specified minimum and maximum length, where the minimum length is $4$ and the maximum length is $30$.
    \item The children of an operator should not all be constants.
    \item The child of a unary operator should not be the inverse of that operator, e.g. $log(exp(x))$.
    \item Descendants of trigonometric operators should not be trigonometric operators, e.g. $sin(x + cos(x))$ is not allowed because cosine is a descendant of since.
\end{enumerate}

\subsubsection{Experiment Setup} Each experiment consists of $2$ million expression evaluations for RSPG, PQT, VPG, and GP. Hyperprameters were tuned using benchmarks Nguyen-$7$ and Nguyen-$10$ using grid search comprising $800$ hyperarameter combinations for GP and $81$ combinations for each of RSPG, PQT, and VPG. Both commercial software algorithms (Eureqa and Wolfram) do not expose hyperparameters and were run until completion. All experiments were replicated with 100 different random seeds for each benchmark expression. Wolfram results are only presented for one-dimensional benchmarks because the method is not applicable to higher dimensions. 

\begin{table*}[h]
\centering

\begin{small}
\begin{tabular}{cccccccc}
Benchmark & Expression & RSPG & PQT & VPG & GP & Eureqa & Wolfram \\ \hline
Nguyen-1 & $x^3+x^2+x$ & 100\% & 100\% & 96\% & 100\% & 100\% & 100\% \\
Nguyen-2 & $x^4+x^3+x^2+x$ & 100\% & 99\% & 47\% & 97\% & 100\% & 100\% \\
Nguyen-3 & $x^5+x^4+x^3+x^2+x$ & 100\% & 86\% & 4\% & 100\% & 95\% & 100\% \\
Nguyen-4 & $x^6+x^5+x^4+x^3+x^2+x$ & 100\% & 93\% & 1\% & 100\% & 70\% & 100\% \\
Nguyen-5 & $\sin(x^2)\cos(x)-1$ & 72\% & 73\% & 5\% & 45\% & 73\% & 2\% \\
Nguyen-6 & $\sin(x)+\sin(x+x^2)$ & 100\% & 98\% & 100\% & 91\% & 100\% & 1\% \\
Nguyen-7 & $\log(x+1)+\log(x^2+1)$ & 35\% & 41\% & 3\% & 0\% & 85\% & 0\% \\
Nguyen-8 & $\sqrt{x}$ & 96\% & 21\% & 5\% & 5\% & 0\% & 71\% \\
Nguyen-9 & $\sin(x)+\sin(y^2)$ & 100\% & 100\% & 100\% & 100\% & 100\% & -- \\
Nguyen-10 & $2\sin(x)\cos(y)$ & 100\% & 91\% & 99\% & 76\% & 64\% & -- \\
Nguyen-11 & $x^y$ & 100\% & 100\% & 100\% & 7\% & 100\% & -- \\
Nguyen-12 & $x^4-x^3+\frac{1}{2}y^2-y$ & 0\% & 0\% & 0\% & 0\% & 0\% & -- \\
\cline{3-8}
 & \multicolumn{1}{r}{Average} & \textbf{83.6\%} & 75.2\% & 46.7\% & 60.1\% & 73.9\% & -- \\ 
\end{tabular}
\end{small}
\caption{Recovery rate comparison of DSO trained using RSPG, PQT, and VPG against three baselines on the Nguyen symbolic regression benchmark suite. A bold value represents statistical significance ($p < 10^{-3}$) across all benchmarks.}
\label{tab:dso_results}
\end{table*}

\begin{figure}
  \centering
  \includegraphics[trim=0 0 0 0, clip, height=1.25in]{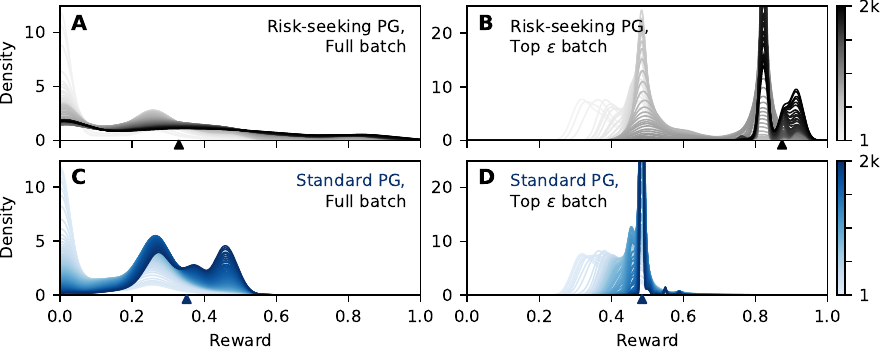}
  \includegraphics[trim=-5 0 0 0, clip, height=1.25in]{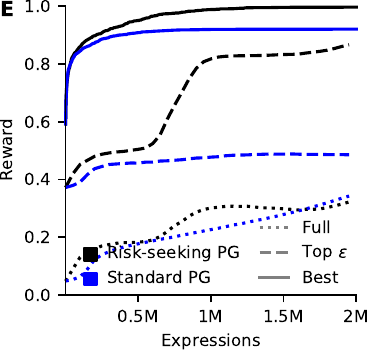}
  \caption{$\textbf{A}$ - $\textbf{D}$. Empirical reward distributions for Nguyen-8. Each curve is a Gaussian kernel density estimate (bandwidth 0.25) of the rewards for a particular training iteration, using either the full batch of expressions ($\textbf{A}$ and $\textbf{C}$) or the top $\epsilon$ fraction of the batch ($\textbf{B}$ and $\textbf{D}$). Black plots ($\textbf{A}$ and $\textbf{B}$) were trained using the risk-seeking policy gradient objective. Blue plots ($\textbf{C}$ and $\textbf{D}$) were trained using the standard policy gradient objective. Colorbars indicate training step. Triangle markings denote the empirical mean of the distribution at the final training step. $\textbf{E}$.Training curves for mean reward of full batch (dotted), mean reward of top $\epsilon$ fraction of the batch (dashed), and best expression found so far (solid), averaged over all training runs.}
  
  \label{fig:distributions}
\end{figure}

\subsubsection{Results and Discussion} 
In \cref{tab:dso_results}, the recovery rate for each benchmark is reported, where exact symbolic equivalence is the definition of recovery. RSPG significantly outperforms PQT, VPG, and all other baselines in its ability to exactly recover benchmark expressions. Importantly, VPG and PQT both outperform GP, Eureqa and Wolfram. 

RSPG explicitly optimizes for best-case performance, possibly at the expense of average performance. This is demonstrated in \cref{fig:distributions} by comparing the empirical reward distributions when trained with either the risk-seeking or standard policy gradient for Nguyen-$8$. Interestingly, at the end of training the mean reward over the full batch (an estimate of $J_{risk} (\theta)$) is larger when training with the standard policy gradient, even though the risk-seeking policy gradient produces larger mean over the top $\epsilon$ fraction of the batch $J_{risk} (\theta; \epsilon)$) and a superior best expression. This is consistant with the intuition of maximizing best-case performance at the expense of average performance. In contrast, the best-case performance of the standard policy gradient plateaus early in training (\cref{fig:distributions}E, dashed blue curve) whereas the risk-seeing policy gradient continues to increase until the end of training (\cref{fig:distributions}E, dashed black curve)

\subsection{Enhancing DSO with Genetic Programming Seeding}
\label{sec:exp:gp}
Next, we evaluate the performance of DSO with genetic programming population seeding described in \cref{sec:gp}. First, we train the proposed DSO and GP hybrid with different objective functions and compare the results to determine the best training method via a number of ablations. Second, we compare the best performing training method against DSO, and benchmark against each objective function described in \cref{sec:training_dso}. For all further experiment and implementation details see \cite{mundhenk2021symbolic}.

\begin{table}[t]
  \centering
  \caption{Recovery rates for various ablations of our algorithm (sorted by overall performance) from 25 independent runs on the Nguyen, R, and Livermore benchmark problem sets.}
    \begin{tabular}{lcccc}
    & \multicolumn{4}{c}{Recovery rate (\%)} \\
    & All & Nguyen & R & Livermore \\
    \cmidrule{2-5}
    Trainer = PQT & \textbf{74.92} & 92.33 & \textbf{33.33} & \textbf{71.09} \\
    Trainer = VPG & 74.27 & \textbf{93.67} & 28.00 & 70.00 \\
    Entropy weight = 0 & 73.95 & 92.00 & 30.67 & 70.00 \\
    Trainer = RSPG & 73.95 & 92.67 & 39.33 & 69.82 \\
    DSO (RSPG only)  & 45.19 & 83.58 & 0.00  & 30.41 \\
    \midrule
    95\% Confidence Interval & $\pm$1.54 & $\pm$1.76 & $\pm$2.81 & $\pm$1.32 \\
    \end{tabular}
  \label{tab:ablations}
\end{table}

\subsubsection{Ablation Studies}
We ran several ablations to determine the contribution of our various design choices to our method's performance. We evaluated our method using the Nguyen and R benchmark datasets. Additionally, we introduce a new benchmark problem set with this work, which we call Livermore. The impetus for introducing a new benchmark problem set was that our algorithm
achieves nearly perfect scores on Nguyen, so we designed a benchmark problem set with a large
range of problem difficulty. The recovery rate for several ablations is presented in \cref{tab:ablations}.

We first note that the choice of RNN training procedure (i.e. VPG, RSPG, or PQT) does not make a large difference; while PQT outperforms VPG and RSPG, the difference falls within the 95\% confidence interval. Interestingly, removing the entropy regularizer by setting the weight to zero made a small difference in performance. Other works have found the entropy regularizer to be extremely important \cite{abolafia2018neural,petersen2019deep,larma2021improving}.
We hypothesize that entropy is less important in our hybrid approach because the GP-produced samples provide the RNN with sufficient exploration. Using different types of mutation in the GP component with equal probability slightly improves performance; we hypothesize that this increases sample diversity. Including the constraints proposed by \cite{petersen2019deep} also improves performance.

\begin{table*}[t]
    \centering
    \caption{Recovery rate of several algorithms on the Nguyen benchmark problem set across 100 independent training runs. Results of our algorithm are obtained using PQT; slightly lower recovery rates were obtained using VPG and RSPG training (see \cref{tab:ablations} for comparisons).
    }
    \begin{tabular}{cccccccc}
    & & \multicolumn{6}{c}{Recovery rate (\%)} \\
    Benchmark & Expression & GP+PQT & RSPG & PQT & VPG & GP & Eureqa \\
    \midrule
    Nguyen-1 & $x^3+x^2+x$ & 100 & 100 & 100 & 96 & 100 & 100 \\
    Nguyen-2 & $x^4+x^3+x^2+x$ & 100 & 100 & 99 & 47 & 97 & 100 \\
    Nguyen-3 & $x^5+x^4+x^3+x^2+x$ & 100 & 100 & 86 & 4 & 100 & 95 \\
    Nguyen-4 & $x^6+x^5+x^4+x^3+x^2+x$ & 100 & 100 & 93 & 1 & 100 & 70 \\
    Nguyen-5 & $\sin(x^2)\cos(x)-1$ & 100 & 72 & 73 & 5 & 45 & 73 \\
    Nguyen-6 & $\sin(x)+\sin(x+x^2)$ & 100 & 100 & 98 & 100 & 91 & 100 \\
    Nguyen-7 & $\log(x+1)+\log(x^2+1)$ & 97 & 35 & 41 & 3 & 0 & 85 \\
    Nguyen-8 & $\sqrt{x}$ & 100 & 96 & 21 & 5 & 5 & 0 \\
    Nguyen-9 & $\sin(x)+\sin(y^2)$ & 100 & 100 & 100 & 100 & 100 & 100 \\
    Nguyen-10 & $2\sin(x)\cos(y)$ & 100 & 100 & 91 & 99 & 76 & 64 \\
    Nguyen-11 & $x^y$ & 100 & 100 & 100 & 100 & 7 & 100 \\
    Nguyen-12 & $x^4-x^3+\frac{1}{2}y^2-y$ & 0 & 0 & 0 & 0 & 0 & 0 \\
    \cmidrule{3-8}
    & \multicolumn{1}{r}{Average} & \textbf{91.4} & 83.6 & 75.2 & 46.7 & 60.1 & 73.9 \\
    \end{tabular}
    \label{tab:results}
\end{table*}

\subsubsection{Results}
Next, we evaluate our DSO GP hybrid against DSO trained with PG, RSPG, and PQT. From the results presented during the ablation, we train the DSO GP hybrid using the PQT algorithm. We follow the experimental procedure of \cref{sec:exp:dso} unless otherwise noted. For all benchmark problems, we run each algorithm multiple times using a different random number seed. 

Our primary empirical performance metric is ``recovery rate,'' defined as the fraction of independent training runs in which an algorithm discovers an expression that is \textit{symbolically equivalent} to the ground truth expression within a maximum of 2 million candidate expressions. The ground truth expression is used to determine whether the best found candidate was correctly recovered. The recovery rates on each of the Nguyen benchmark problems compared with DSO, PQT, VPG, GP, and Eureqa \cite{schmidt2009distilling} are shown in \cref{tab:results}.
We note that RSPG stands as the prior top performer on this set. As we can see, the recovery rate for our algorithm is 9.3\% higher than the previous leader, RSPG. Therefore, seeding GP populations with solutions generated by the RNN and training via PQT ensures solutions of higher fitness are recovered when compared to training the RNN without additional evolved samples from a GP.

\subsection{Integrating Pre-Trained Priors into DSO}
\label{sec:exp:lspt}

In this section, we evaluate the impact of using 
a pre-trained prior (PP) on the performance of DSO. 

\subsubsection{Dataset Overview and Statistics} 
From a raw Wikipedia dump file of $\sim$70 GB, we extracted 798,998 mathematical expressions across 41,763 different pages. As an example of using the category hierarchies, when we search pages under the category \textit{Physics} with depth 3, we collect 67,404 expressions from 2,265 pages in 879 categories, while \textit{Physics} itself has 25 pages with 1,374 expressions.

\begin{table*}
\centering
\caption{Comparison of recovery rate, mean steps to solve, and mean rate of invalid expressions in the symbolic regression task with and without the PP.}
\begin{tabular}{cccc|ccc}
 & \multicolumn{3}{c}{DSO without PP}  & \multicolumn{3}{c}{DSO with PP}  \\
      Benchmark & Recovery & Steps & Invalid       & Recovery & Steps & Invalid  \\ \hline
$x^3+x^2+x$ & 100.\% & 165.2 & 47.82 & 100.\% & 99.47 & 56.75 \\ 
$x^4+x^3+x^2+x$ & 100.\% & 264.6 & 35.86 & 100.\% & 235.6 & 36.46 \\ 
$x^5+x^4+x^3+x^2+x$ & 100.\% & 349.2 & 28.35 & 100.\% & 329.1 & 26.93 \\ 
$x^6+x^5+x^4+x^3+x^2+x$ & 100.\% & 672.2 & 17.65 & 100.\% & 525.5 & 20.75 \\ 
$\sin(x^2)\cos(x)-1$ & 76.\% & 847.8 & 23.67 & 94.\% & 672.8 & 22.81 \\ 
$\sin(x)+\sin(x+x^2)$ & 100.\% & 189.3 & 45.26 & 100.\% & 120.3 & 51.63 \\ 
$\log(x+1)+\log(x^2+1)$ & 35.\% & 1513. & 11.02 & 27.\% & 1620. & 10.91 \\ 
$\sqrt{x}$ & 95.\% & 601. & 31.81 & 99.\% & 365.1 & 32.62 \\ 
$\sin(x)+\sin(y^2)$ & 100.\% & 117.8 & 34.35 & 100.\% & 95.52 & 27.34 \\ 
$2\sin(x)\cos(y)$ & 100.\% & 368.3 & 17.55 & 100.\% & 364.3 & 13.93 \\ 
$x^{y}$ & 100.\% & 22.36 & 56.37 & 100.\% & 12.58 & 41.13 \\ 
$x^4-x^3+\frac{1}{2}y^2-y$ & 0.\% & 2000. & 6.373 & 0.\% & 2000. & 5.71 \\ 
\cline{2-7}
\multicolumn{1}{r}{Average:} & 83.8\% & 592.6 & 29.7\% & \textbf{85.0\%} & \textbf{536.7} & \textbf{28.9\%} 
\end{tabular}
\label{tab:sr-results}
\end{table*}

\subsubsection{Experiment Setup} 
We demonstrate the value of the PP by using it to inform the task of symbolic regression. For simplicity, we replicate the experimental setup and hyperparameters detailed in \cref{sec:exp:dso}. The only change is the introduction of the pre-trained prior (PP) as previously described, sweeping over inverse temperature hyperparameter $\lambda \in \{0.1, 0.2, \dots, 1.0\}$.
The PP is trained for 200 epochs using a recurrent neural network with a hidden layer size of 256, showing the cross-entropy loss going down to 0.367.

\subsubsection{Results}  
In \cref{tab:sr-results}, we report the recovery rate (fraction of runs in which the exact symbolic expression is found), average number of steps required to find the solution, and the fraction of invalid expressions (those that produce floating-point errors, e.g. overflows) produced during training, over 100 independent runs.

The PP show an improvement in recovery rate, requiring fewer steps and generating fewer invalid expressions. In particular, we note the dramatic reduction in steps required to find expressions $\sqrt{x}$, $\sin(x)+\sin(y^2)$, and $x^{y}$. 

Overall, these results show that the use of the PP provides a better informed search in symbolic regression tasks.

\subsection{Unified Framework: Synergistic Integration of DSO Components}
\label{sec:exp:udso}

Here, we rewire each solution strategy defined in this work and create an algorithmic framework that leverages their key capabilities. We evaluate and compare our unified approach against many contemporary SR methods. Finally, we perform a number of ablations on our framework to assess the relative contributions of each component in various contexts.

\subsubsection{Unified DSO}
For our unified DSO (uDSO) framework, we combine individual modules presented in \cref{sec:dso} and \cref{sec:training_dso}. Therefore, the autoregressive model, large scale pre-training, genetic programming population seeding, and constant optimization are combined as overlapping but independent modules from which solutions of high fitness can be generated.

\begin{figure}[t]
    \centering
    \includegraphics[scale=0.6]{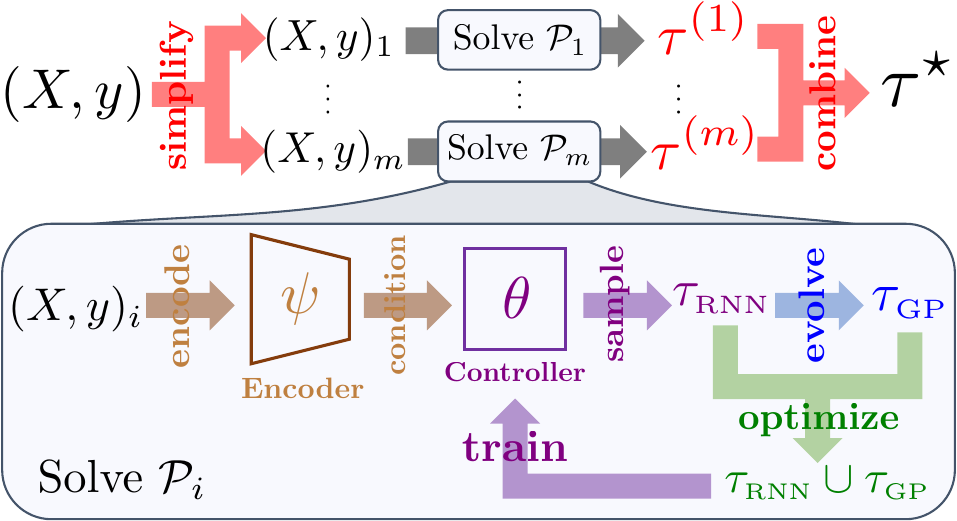}
    \caption{\textbf{Unified deep symbolic optimization.} The five integrated solution strategies are color-coded: \textcolor{aif}{AIF}, \textcolor{dsr}{DSO}, \textcolor{lspt}{LSPT}, \textcolor{gp}{GP}, \textcolor{lm}{LM}. 
    The solution process begins with an initial offline stage, where a parametric controller is pre-trained on a large dataset (LSPT). In the online stage, a recursive problem simplification module (AIF) produces sub-problems of lower dimensionality for the main trunk of uDSO). Within the trunk, an \ar model (RNN in this case) continually learns over many iterations (DSO) and provides good candidates to seed the starting population for genetic programming (GP); high fitness populations produced by GP are combined with elite candidates for controller parameter updates via risk-seeking policy gradient. Finally, the RNN is permitted the use of a powerful symbolic token that encapsulates an entire space spanned by basis functions (LM).
    }
    \label{fig:udso_enter-label}
\end{figure}

The uDSO optimization process starts with an initial offline stage, where a parametric controller is pre-trained
model on a large dataset (LSPT). This step facilitates generalization to test problem instances. Within the trunk, a Recurrent Neural Network (RNN) continually learns over many iterations (DSO) and provides good candidates to seed the starting population for genetic programming (GP); high fitness populations produced by GP are combined with elite candidates for controller parameter updates via risk-seeking policy gradient. Finally, the RNN is permitted the use of a powerful symbolic token that encapsulates an entire space spanned by basis functions (LM). An additional problem simplification step is also included in the uDSO framework. AI Feynman (AIF) \cite{udrescu2020ai} is used to recursively simplify a problem, $P$, to $m$ sub-problems, where each sub-problem, $P_{i}$, is individually solved using the various components of the uDSO framework. Using AIF the solutions of each sub-problem, $\tau^{i}, ..., \tau^{m}$, are then combined into a final solution, $\tau^{*}$. The uDSO framework is illustrated in \cref{fig:udso_enter-label}. For all implementation details and hyperparameters we refer the reader to the work of Landajuela et al. \cite{landajuela2022unified}.

\subsubsection{Pre-Training Strategies in the Unified Framework}
We pre-train four models: two trained using supervised learning and two trained using reinforcement learning, each with and without the \Tpoly token. Details of the pre-training setup can be found in \cref{sec:pretraining_encoder_decoder}.

\subsubsection{Experiment Setup}
We empirically assess uDSO using SRBench \cite{la2021contemporary}, by running uDSO through the SRBench pipeline, we enable direct comparison to its curated results of 14 contemporary SR methods.

For each experiments, we use a minimal set of tokens: \Tadd, \Tsub, \Tmul, \Tdiv, \Tsin, \Tcos, \Texp, \Tlog, \Tsqrt, \Tone, \Tconst, and \Tpoly (except for the appropriate ablations).
For simplicity, our choice of $\Phi$ (basis functions for \Tpoly) includes only polynomial terms up to degree 3. When possible, we reused the hyperparameters of the published methods most similar to each uDSO component.

\subsubsection{Results and Analysis}

\begin{figure}[t]
    \centering
    \includegraphics[width=0.495\linewidth]{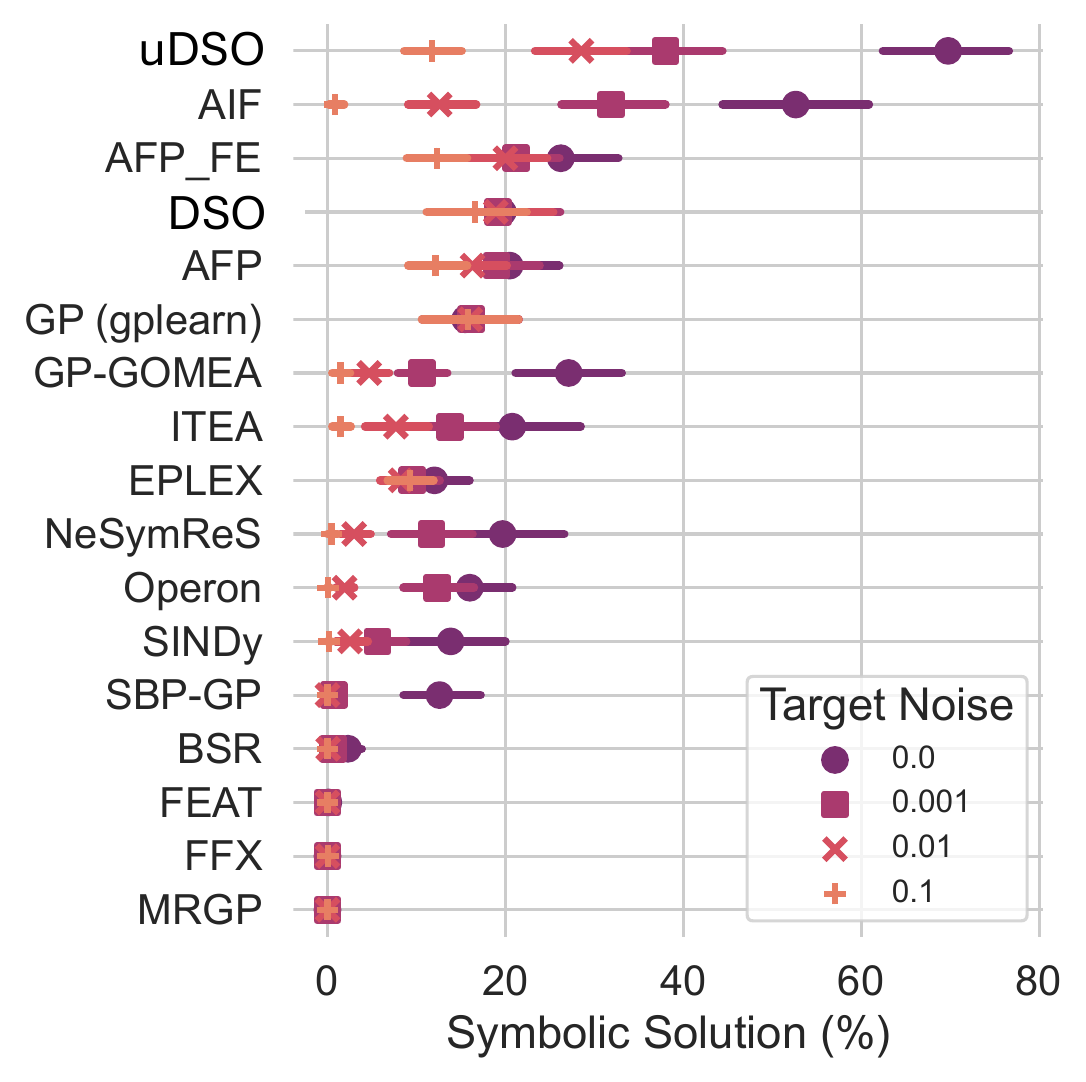}
    \includegraphics[width=0.495\linewidth]{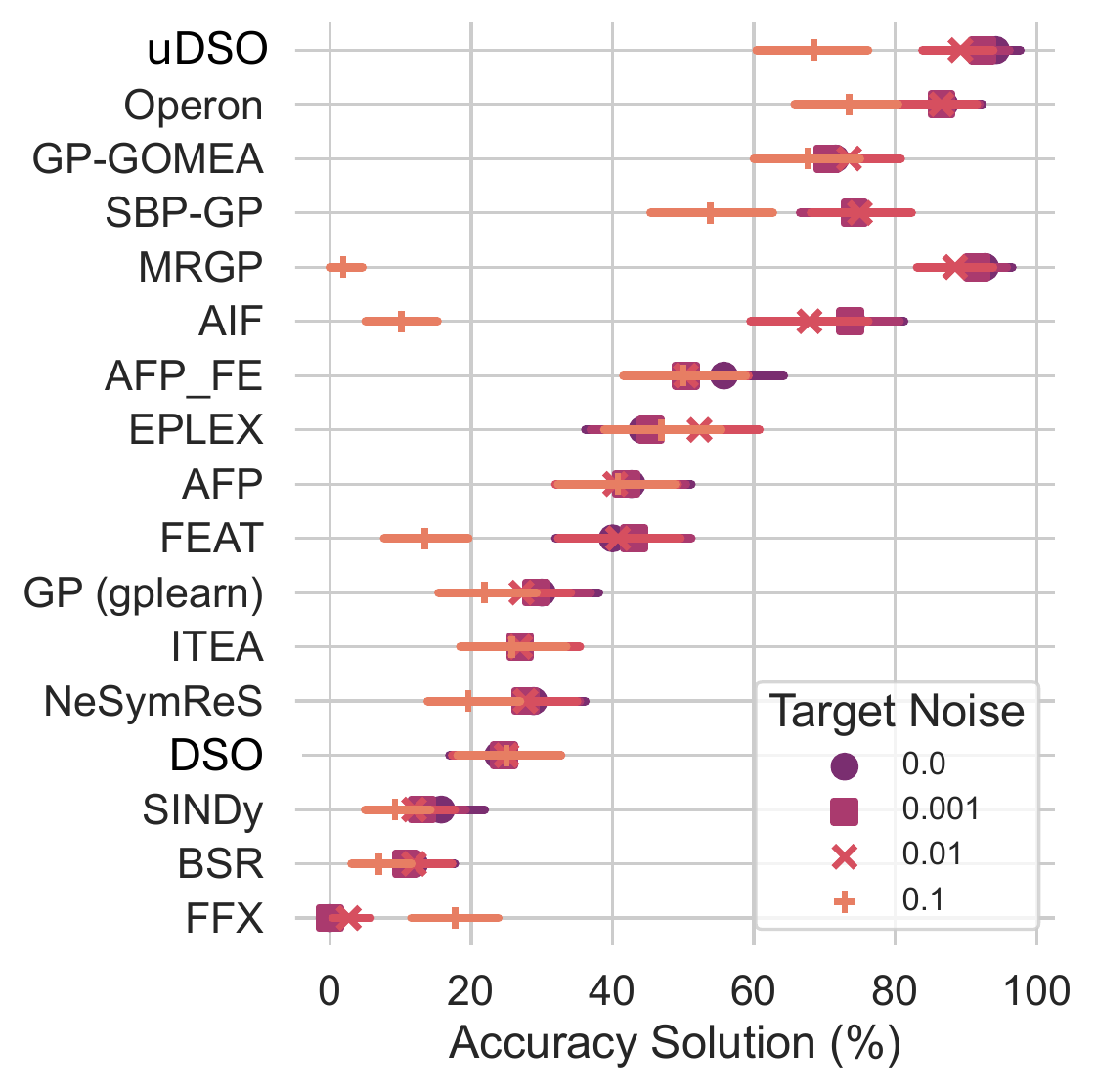}
    \caption{SRBench-generated comparisons of symbolic (left) and accuracy-based (right) solution rates for uDSO and 16 baseline SR methods, averaged across 130 ground-truth SR problems and four different noise levels.}
    \label{fig:ground-truth}
\end{figure}

\begin{figure}
    \centering
    \includegraphics[width=0.6\linewidth]{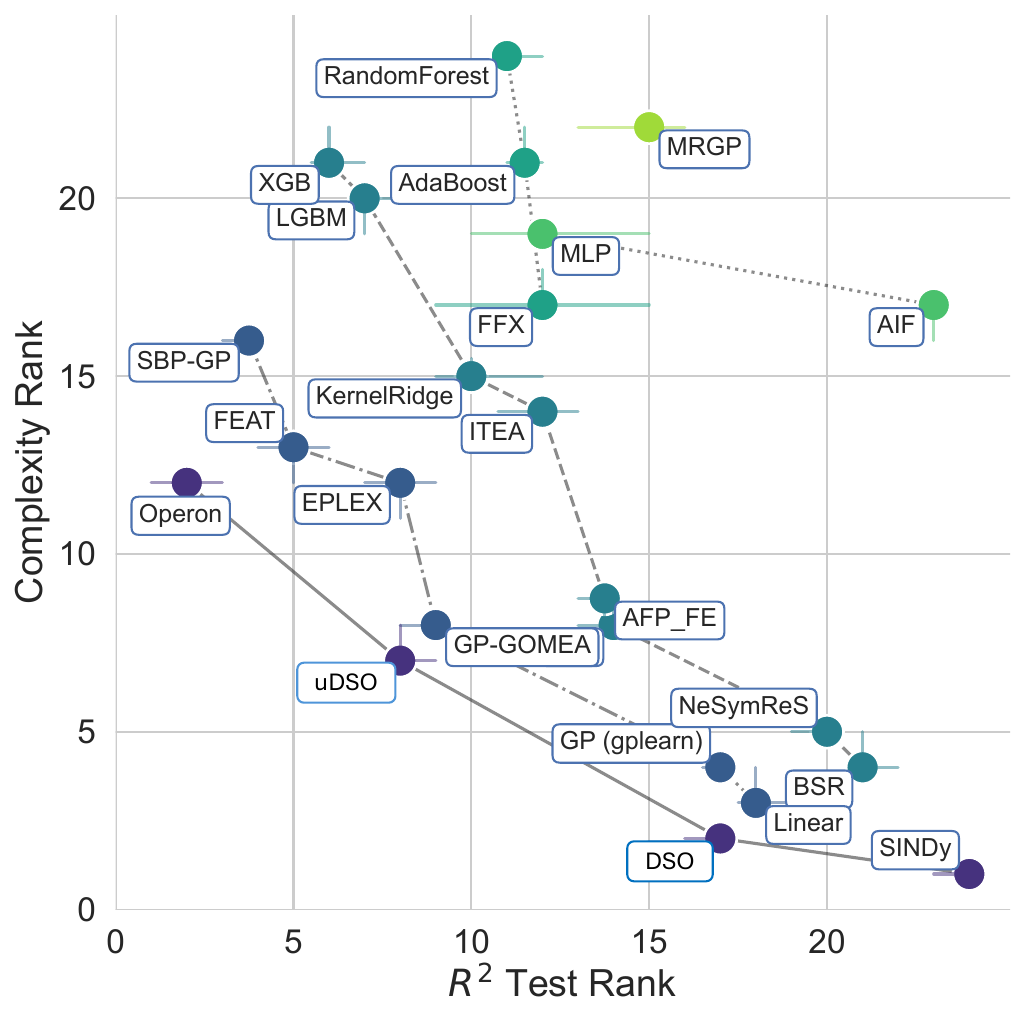}
    \caption{SRBench-generated comparison of test $R^2$ (on held-out test data) vs complexity for uDSO and 22 baseline regression/SR methods, averaged across 122 black-box SR problems.}
    \label{fig:black-box}
\end{figure}

As shown by \cref{fig:ground-truth}, uDSO outperforms all other 14 benchmarked methods both in symbolic solution (by a large margin) and accuracy solution rates. The Pareto frontier is presented in \cref{fig:black-box}, and shows that uDSO falls on the Pareto frontier, alongside Operon \cite{kommenda2020parameter} (at higher accuracy and complexity), and DSO and SINDy (at lower accuracy and complexity).

Notably, the previously published SRBench results had no clear winner across the three categories of (1) symbolic solution, (2) accuracy solution, and (3) Pareto efficiency. The top performers for symbolic solution rate (namely, AIF) differed from those for accuracy solution rate (namely, Operon and MRGP). Only Operon appeared as a top performer in two categories. However, in this work, we see that uDSO is highest in all three categories.

\subsubsection{Component Ablation Studies}
Since uDSO integrates many disparate components, it is critical to assess the relative contributions of each component in various contexts.
To identify synergies, anti-synergies, and diminishing returns among the integrated methods, we perform a large-scale study of \textit{all combinations} of the five integrated methods
(standard ``leave-one-out'' style ablations in which we begin with the full uDSO and ablate one component at a time do not provide this full picture). 

There are 48 total combinations: AIF $\in$ [on, off] $\times$ GP $\in$ [on, off] $\times$ \Tpoly $\in$ [on, off] $\times$ pre-training $\in$ [SL, RL, off] $\times$ DSO $\in$ [on, off].
We define ablating each component as follows:
For AIF, ``off'' means that we do not perform recursive problem-simplification; only the root problem is considered.
For GP, ``off'' means we do not perform inner-loop GP algorithms between batches.
For \Tpoly, ``off'' means we exclude the \Tpoly token from the library.
For pre-training, SL and RL refer to whether we use the model pre-trained using SL or RL; ``off'' means we do not use any pre-trained model, and the controller architecture does not include the set transformer component.
For DSO, ``off'' means the learning rate for the neural network is set to zero; notably, the controller is still used for sampling, including all priors and constraints.
For each ablation, we run on the 130 ground-truth SRBench problems with 10 random seeds each ($48 \times 130 \times 10 = 62,400$ total runs), using a maximum of 500,000 expression evaluations per sub-problem.

\begin{figure}
    \centering
    \includegraphics[width=1.0\linewidth]{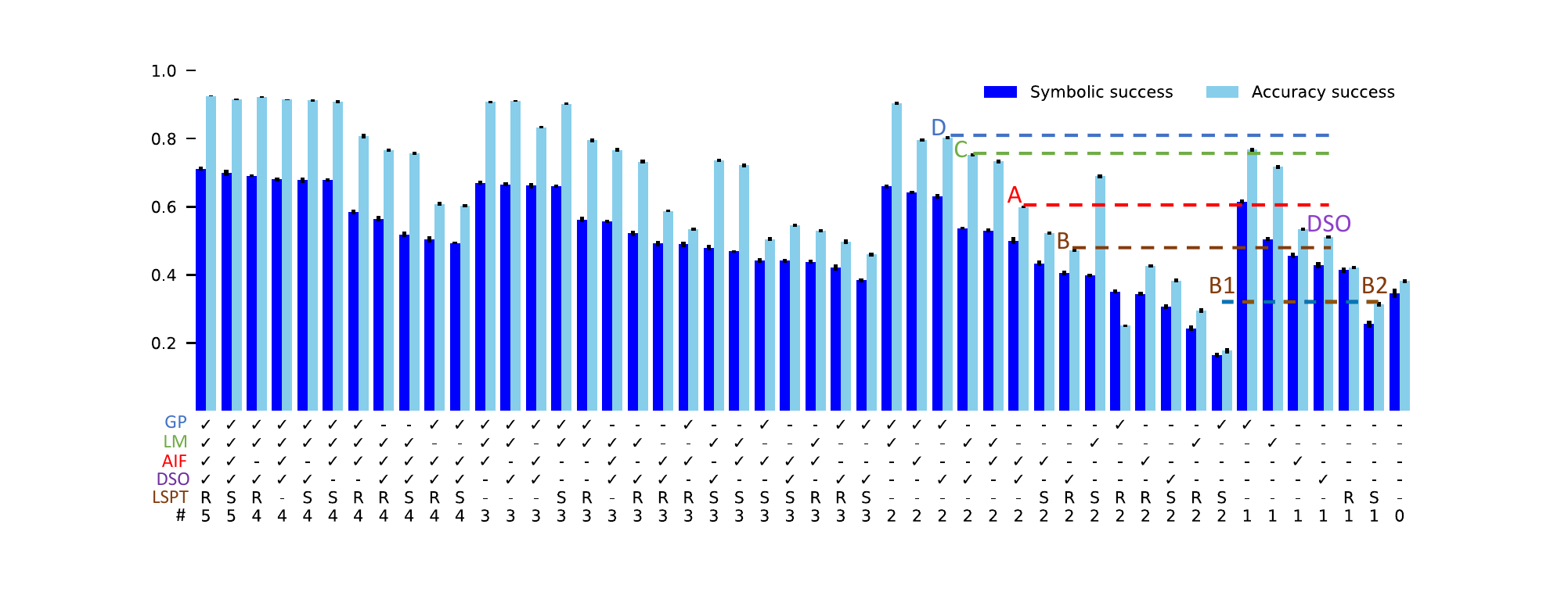}
    \caption{Combinatorial ablations of uDSO components on 130 SRBench ground-truth problems.
    Indicators below bars denote which components are enabled for each ablation.
    \checkmark: Component enabled.
    -: Component disabled.
    S: Pre-training enabled using SL.
    R: Pre-training enabled using RL.
    \#: Number of enabled components.
    Error bars represent standard error across 10 random seeds per problem.
    Colored labels are referred to in the main text.
    Ablation `D' is equivalent to \cite{mundhenk2021symbolic}.
    Ablation `DSO' is equivalent to \cite{larma2021improving}.}
    \label{fig:ablations}
    \vspace{-12pt}
\end{figure}

The symbolic solution rate and accuracy solution rate computed by SRBench is presented in \cref{fig:ablations}. Additionally, labels for select ablations are also presented in \cref{fig:ablations}. Generally, performance increases with the number of enabled components, $N$. For a fixed $N$, GP and LM tend to yield the largest marginal performance gains.

%% file: conclusion.tex
\section{Conclusions}
In this chapter, we present a novel computational framework for scientific discovery called deep symbolic optimization (DSO). The core elements of DSO include an autoregressive model which is used to generate solutions. By sampling in an autoregressive manner it is possible to incorporate domain knowledge via in-situ priors and constraints. We build upon the core elements of the DSO framework with additional components by incorporating reinforcement learning objective functions, evolutionary methods, and large-scale pre-training. We evaluate each DSO component using the symbolic regression problem domain and compare our performance against various baselines from the literature. Our evaluation procedure culminates in a unified comparison of all DSO components by combining each component to generate a unified DSO (uDSO) algorithm. Using this approach, we compare against relevant contemporary SR algorithms, and show that uDSO achieves state-of-the-art performance. By applying DSO to SR we show that DSO can generate interpretable solutions for each benchmark dataset. DSO not only generates highly fit solutions but solutions that are predictable, interpretable, and transparent, which is an additional advantage over many other deep learning approaches.

We evaluate DSO using the SR problem domain. However, the DSO framework has been applied in many other problem domains relevant for scientific discovery. DSO has also been applied to power converter design \cite{glatt2021deep} and symbolic policy discovery for reinforcement learning problems \cite{landajueladiscovering}. DSO has been combined with large language models to accelerate learning in the antibody optimization domain and to learn sparse policies for sepsis treatment \cite{pettit2021learning}. More recently, DSO has been applied to antibody optimization, where the framework has been expanded for multi-objective \cite{faris2024pareto} and multi-fidelity \cite{dasilva2023multi_fidelity} settings.

In the future, we aim to further extend the multi-objective capabilities by including advanced multi-objective evolutionary algorithms (MO-EA) \cite{zitzler1999multiobjective}, e.g. NSGA-II \cite{deb2002fast}, into the DSO framework. This will enable multi-objective DSO methods to bootstrap from the evolved solutions computed by the MO-EAs. We also aim to include autoregressive pre-pretrained large language models (LLMs) \cite{wolf2019huggingface}
in the DSO framework and fine-tune them online for a given task using reinforcement learning. By including pre-trained autoregressive LLMs in the DSO framework, we can exploit the extensive \emph{a priori} knowledge available and potentially speed up scientific discovery.

\section*{Acknowledgements}
We thank Livermore Computing at Lawrence Livermore National Laboratory (LLNL) for the computational resources that enabled this work. Funding was provided by the LLNL Laboratory Directed Research and Development projects 19-DR-003 and 21-SI-001.
We thank the Computational Engineering Directorate and the Data Science Institute at LLNL for additional support.
This work was performed under the auspices of the U.S. Department of Energy by LLNL under contract DE-AC52-07NA27344.
LLNL-JRNL-863241.